\documentclass[12pt,a4paper]{article}

\usepackage[margin=1in]{geometry}
\usepackage{times}
\usepackage{setspace}
\usepackage{amsmath,amssymb,amsthm,mathtools,bm}
\usepackage{booktabs,array,multirow,graphicx}
\usepackage{placeins}
\usepackage{enumitem}
\usepackage{algorithm}
\usepackage{algpseudocode}
\usepackage{xcolor}
\usepackage{microtype}
\usepackage{natbib}
\usepackage[hidelinks]{hyperref}
\usepackage{aliascnt}
\usepackage[nameinlink,capitalise,noabbrev]{cleveref}
\usepackage{url}

\allowdisplaybreaks
\numberwithin{equation}{section}

\newtheorem{theorem}{Theorem}[section]
\newaliascnt{proposition}{theorem}
\newtheorem{proposition}[proposition]{Proposition}
\aliascntresetthe{proposition}
\newaliascnt{lemma}{theorem}
\newtheorem{lemma}[lemma]{Lemma}
\aliascntresetthe{lemma}
\newaliascnt{corollary}{theorem}
\newtheorem{corollary}[corollary]{Corollary}
\aliascntresetthe{corollary}
\newaliascnt{assumption}{theorem}
\newtheorem{assumption}[assumption]{Assumption}
\aliascntresetthe{assumption}
\crefname{assumption}{Assumption}{Assumptions}
\Crefname{assumption}{Assumption}{Assumptions}
\theoremstyle{definition}
\newaliascnt{definition}{theorem}

\aliascntresetthe{definition}
\newaliascnt{remark}{theorem}

\aliascntresetthe{remark}
\newaliascnt{example}{theorem}

\aliascntresetthe{example}

\crefname{proposition}{Proposition}{Propositions}
\Crefname{proposition}{Proposition}{Propositions}
\crefname{lemma}{Lemma}{Lemmas}
\Crefname{lemma}{Lemma}{Lemmas}
\crefname{corollary}{Corollary}{Corollaries}
\Crefname{corollary}{Corollary}{Corollaries}

\crefname{algorithm}{Algorithm}{Algorithms}
\Crefname{algorithm}{Algorithm}{Algorithms}
\algrenewcommand{\algorithmicrequire}{\textbf{Input:}}
\algrenewcommand{\algorithmicensure}{\textbf{Output:}}

\newcommand{\E}{\mathbb E}
\newcommand{\Pp}{\mathbb P}
\newcommand{\1}{\mathbf 1}
\newcommand{\cF}{\mathcal F}
\newcommand{\cI}{\mathcal I}
\newcommand{\cE}{\mathcal E}
\newcommand{\cB}{\mathcal B}
\newcommand{\cD}{\mathcal D}
\newcommand{\cN}{\mathcal N}
\newcommand{\eps}{\varepsilon}
\newcommand{\wh}{\widehat}
\newcommand{\wt}{\widetilde}
\newcommand{\ol}{\overline}
\newcommand{\ul}{\underline}
\newcommand{\Var}{\operatorname{Var}}
\newcommand{\Cov}{\operatorname{Cov}}

\title{Infinite Quantiles Inference under Local Differential Privacy}
\author{Yi Liu\thanks{Department of Mathematics and Statistics, York University}\hspace{0.4em} and 
Qirui Hu \thanks{\hangindent=1.8em\hangafter=1 School of Statistics and Data Science, Shanghai University of Finance and Economics.\\
Department of Mathematics, Ruhr University Bochum} \thanks{Corresponding author: huqirui@mail.shufe.edu.cn}\hspace{0.6em}}
\date{}

\begin{document}
\maketitle

\begin{abstract}
We study simultaneous quantile inference under local differential privacy for a number of target probabilities that grows with the sample size. Each user participates once, so learning unknown quantile locations and calibrating their joint uncertainty must share the same private information. We fuse complementary cumulative distribution function (CDF) scores from a single tree report using their exact covariance; the variance at the correct anchor is no larger than that of either score, and anchor error incurs only quadratic excess variance. Three consecutive cohorts then localize the quantiles, estimate their densities, and supply fresh scores for a one-step correction and joint multiplier calibration. Under mild conditions, a uniform representation controls localization and density errors and yields asymptotically valid simultaneous coverage. A fixed-privacy regime permits every polynomially growing target grid. Simulations show consistent same-report precision gains, while comparisons with CDF-band inversion expose the sample cost of learning target locations. Census microdata illustrates how the resulting joint intervals support wage-percentile ratios.
\end{abstract}


\clearpage
\section{Introduction}\label{sec:intro}

From 1987 to 2004, U.S. upper-half wage inequality rose while lower-half inequality contracted \citep{AutorKatzKearney2006}. Quantile comparisons distinguish these distributional changes that a mean or an isolated percentile can miss. As a data stream grows, a finer percentile grid permits more detailed comparisons. Their uncertainty must be assessed jointly: a collection of pointwise intervals does not control the largest error over an expanding grid. We study this problem when each user privatizes a measurement before sending it to the server and participates only once.

Local differential privacy (LDP) makes this joint inference problem different from classical quantile inference. Sample-quantile representations and cumulative distribution function (CDF) bands connect nonprivate quantile error to a shared empirical process \citep{Bahadur1966,Csorgo1983,Massart1990,CFM2013}. Under LDP, the server receives randomized reports rather than observations \citep{Warner1965,Duchi2018}. It must learn where the quantiles are while deciding which thresholds to query. Giving each percentile a separate user group loses information as the number of targets grows; using a shared report avoids this split but creates dependent errors at adaptively selected thresholds. Joint calibration must accommodate both that dependence and the error in the learned locations.

Recent work provides online inference for a prespecified scalar quantile. \citet{Liu2023} use randomized binary inquiries and self-normalization to avoid nuisance-density estimation. \citet{Cai2025} use dynamically growing parallel stochastic-approximation paths to obtain an asymptotic confidence sequence over time. Other developments address communication-efficient single-quantile estimation, federated scalar inference, and fixed-dimensional quantile-regression contrasts \citep{Aamand2025,Cai2026Fed,Cai2026Integration,LiuHu2026QR}. Our target is joint inference at a terminal horizon fixed before data collection, with the number of probability targets increasing as that horizon grows. Extending scalar inference to this setting requires uniform control of localization and density errors at the scale of a maximum, rather than a separate scalar approximation at every target.

CDF estimation followed by inversion is an important alternative. \citet{Liu2024CDF} study private CDF estimation on public inspection grids. A simultaneous CDF band supports probability queries selected after release, whereas direct inference learns locations for prespecified probabilities.

For independent observations with CDF $F$, our targets are
\[
 0<\tau_{1,n}<\cdots<\tau_{k_n,n}<1,
 \qquad Q_{j,n}=F^{-1}(\tau_{j,n}),
\]
where the deterministic number $k_n$ may diverge. ``Infinite'' refers to this increasing sequence of finite target grids. With one report per user at privacy budget $\eps_n>0$ and public queries allowed to depend on past reports, we seek intervals satisfying
\[
 \Pp\{Q_{j,n}\in\cI_{n,j}\text{ for every }j\le k_n\}
 \longrightarrow 1-\alpha,
 \qquad \alpha\in(0,1).
\]
The probability targets are specified before collection; the method learns their spatial locations.

Our first step extracts more information from a shared range-tree report with optimized unary encoding (OUE) \citep{Wang2017,Cormode2019}. Every threshold has two unbiased CDF scores: a prefix sum and the complement of a suffix sum. Their exact covariance determines a minimum-variance linear fusion at the correct CDF anchor. Fusion remains unbiased for any public anchor, and the variance penalty for an inaccurate anchor is quadratic. Thus the target probability itself supplies an increasingly accurate anchor as localization improves. This gain changes only decoding, with no additional report or privacy expenditure.

Our second step separates adaptive learning from joint calibration. The first cohort averages one projected stochastic-approximation path for each target. The second estimates densities near the learned locations. The third supplies fresh user-level score vectors. A one-step correction cancels first-order localization error, leaving a product of localization and density errors and a local nonlinear remainder. Conditional on the first two cohorts, the fresh vectors are independent and identically distributed; shared Gaussian multipliers retain their cross-quantile dependence. We bound the remainder uniformly and connect it to high-dimensional Gaussian approximation \citep{PolyakJuditsky1992,CCK2013,CCK2017}. At fixed privacy, a sufficient regime permits every polynomial growth rate of $k_n$, with only logarithmic grid-size factors in the width bound.

We compare a same-channel fixed-grid inversion procedure with the complete three-cohort method, including the users spent on learning. Fusion reduces error by 21--23\% relative to balanced decoding in the matched three-stage experiments. CDF inversion gives smaller errors and narrower, more conservative intervals on the tested public spatial grids. These comparisons distinguish the benefit of extracting complementary information from a report from the cost of learning where to query it.

The paper is organized as follows. Section~\ref{sec:method} develops the procedure and its implementation. Section~\ref{sec:theory} states the assumptions and theoretical guarantees, and Section~\ref{app:fixed-results} compares fixed-chain calibration. Sections~\ref{sec:simulation} and~\ref{sec:application} present the complete numerical experiments and wage-percentile application. Section~\ref{sec:conclusion} concludes. Proofs and their auxiliary lemmas are collected in the appendices.

\section{Shared private scores and quantile inference}\label{sec:method}

\Cref{alg:main} gives the complete procedure, from public inputs to simultaneous intervals.  We first construct the shared private report and then describe the three cohorts and default tuning; \Cref{sec:streaming} gives the streaming updates.

\label{sec:setup}

We suppress the subscript $n$ on $\tau_{j,n}$ and $Q_{j,n}$ when the sample size is clear.  The generalized inverse is
\[
 F^{-1}(u)=\inf\{x\in\mathbb R:F(x)\ge u\}.
\]
The public projection bracket $[L,U]$ contains all target quantiles in its
interior, while the raw observations may be unbounded. Public scientific
range information or an independently privatized range-estimation cohort
can supply this bracket. The Stage-1 projection frequency and the distance
of final iterates from its endpoints summarize its adequacy. At time $i$, a
public query $\mathsf q_i$ may be a measurable function of the previous
reports $Z_1,\ldots,Z_{i-1}$.  A local randomizer $\mathcal M_i$ is
$\eps_n$-LDP conditional on the public history if, for every pair $x,x'$,
every history, and every measurable report set $A$,
\[
 \Pp\{\mathcal M_i(x;\mathsf q_i)\in A\mid Z_{1:i-1}\}
 \le e^{\eps_n}
 \Pp\{\mathcal M_i(x';\mathsf q_i)\in A\mid Z_{1:i-1}\}.
\]
The complete procedure is user-level $\eps_n$-LDP for every deterministic horizon.  Conditional on the past transcript, OUE has this privacy bound and the sampled tree level is independent of the user's value.  History-dependent queries and server-side computations preserve the bound; each person belongs to only one cohort.  \Cref{app:privacy} gives the likelihood-ratio argument for the full sequential transcript.

\subsection{Complementary scores from one report}\label{sec:tree}

For privacy budget $\eps=\eps_n$, fix $d$ ordered public thresholds
\[
 x_{(1)}\le\cdots\le x_{(d)}.
\]
Stable tags distinguish ties.  These thresholds induce $d+1$ interval categories
\[
 C_0=(-\infty,x_{(1)}],\quad
 C_s=(x_{(s)},x_{(s+1)}]\ (1\le s<d),\quad
 C_d=(x_{(d)},\infty).
\]
Let $H_d=\lceil\log_2(d+1)\rceil$ and pad the $d+1$ real leaves by zero-probability leaves to obtain a complete binary tree with $2^{H_d}$ leaves.  Levels are numbered $1,\ldots,H_d$, with the root excluded.  A user with value $X$ determines the leaf $C(X)$, draws a level $L^\star$ uniformly from $\{1,\ldots,H_d\}$, and finds the unique node on the path to $C(X)$ at that level \citep{Cormode2019}.  The node is represented by a unary vector.  With $
 p_\eps=1/2,
 q_\eps=1/(e^\eps+1), $
OUE \citep{Wang2017} independently releases every bit as one with probability $p_\eps$ when its input bit is one and with probability $q_\eps$ when its input bit is zero.  The report consists of the sampled level and the privatized unary vector.

For a node $v$ at level $\ell(v)$, let $Z_v$ be its released bit, or zero on an unsampled level, and define
\[
 \psi_v(Z)
 =H_d\1\{L^\star=\ell(v)\}
   \frac{Z_v-q_\eps}{p_\eps-q_\eps}.
\]
The privacy-noise constant and a deterministic score envelope are
\begin{equation}\label{eq:nu-envelope}
 \nu_\eps=\frac{4e^\eps}{(e^\eps-1)^2},
 \qquad
 \cB_{d,\eps}=1+\frac{2H_de^\eps}{e^\eps-1}.
\end{equation}

For threshold rank $r\in\{1,\ldots,d\}$, let $\cD_r^-$ be the canonical disjoint-node decomposition of the first $r$ leaves and let $\cD_r^+$ be the decomposition of the complementary leaves.  Each decomposition consists of the maximal tree nodes whose leaves lie entirely in the required set.  Write $w_r^-=|\cD_r^-|$ and $w_r^+=|\cD_r^+|$; both are between one and $H_d$, and each decomposition contains at most one node at any level.  Define
\[
 A_r=\sum_{v\in\cD_r^-}\psi_v,
 \qquad
 B_r=1-\sum_{v\in\cD_r^+}\psi_v.
\]
Both estimate $F(x_{(r)})$.  We choose the fusion weight by
evaluating the variances and covariance of $A_r$ and $B_r$ at a public reference
value $u\in(0,1)$ for $F(x_{(r)})$.  For this anchor, set
\begin{align*}
 a_r^-(u)&=H_dw_r^-\nu_\eps+(2H_d-1)u,\\
 a_r^+(u)&=H_dw_r^+\nu_\eps+(2H_d-1)(1-u),\\
 \lambda_r^\circ(u)&=\frac{a_r^+(u)}{a_r^-(u)+a_r^+(u)},
\end{align*}
and form the fused score
\begin{equation}\label{eq:fused-score}
 G_r(u)=\lambda_r^\circ(u)A_r+\{1-\lambda_r^\circ(u)\}B_r.
\end{equation}

\begin{proposition}[Exact fused-score moments]\label{prop:fused-moments}
Let $p_r=F(x_{(r)})$.  Condition on the pre-report public history,
including the threshold list and the anchor $u$, so both are fixed.  Then
\begin{align*}
 \E A_r=\E B_r&=p_r,\quad
 \Cov(A_r,B_r)=p_r(1-p_r)\\
 \Var(A_r)&=H_dw_r^-\nu_\eps+2H_dp_r-p_r^2,\\
 \Var(B_r)&=H_dw_r^+\nu_\eps+2H_d(1-p_r)-(1-p_r)^2.
\end{align*}
Consequently, $G_r(u)$ is unbiased for every anchor fixed by this
history.  At $u=p_r$,
\begin{equation}\label{eq:fused-min-var}
 \Var\{G_r(p_r)\}
 =p_r(1-p_r)+
 \frac{a_r^-(p_r)a_r^+(p_r)}{a_r^-(p_r)+a_r^+(p_r)}
 \le \min\{\Var(A_r),\Var(B_r)\}.
\end{equation}
For an arbitrary anchor $u$,
\begin{equation}\label{eq:fusion-excess}
 \Var\{G_r(u)\}-\Var\{G_r(p_r)\}
 =\{a_r^-(p_r)+a_r^+(p_r)\}
   \{\lambda_r^\circ(u)-\lambda_r^\circ(p_r)\}^2,
\end{equation}
with the explicit bound
\begin{equation}\label{eq:fusion-excess-bound}
 0\le \Var\{G_r(u)\}-\Var\{G_r(p_r)\}
 \le (2H_d-1)(u-p_r)^2.
\end{equation}
Moreover, $|G_r(u)|\le\cB_{d,\eps}$ almost surely.
\end{proposition}

At the correct anchor, fusion minimizes variance among linear combinations with weights summing to one. The shorter-side balanced decoder instead chooses $A_r$ or $B_r$ by node count. All stages use $u=\tau_j$ for target $j$, including both density-estimation thresholds. Unbiasedness holds throughout adaptation; \eqref{eq:fusion-excess-bound} makes the variance cost vanish quadratically as the queried CDF value approaches $\tau_j$.

\subsection{Adaptive localization}\label{sec:stage1}

The total horizon $n$ and stage fractions are fixed before data collection.  Set
\begin{equation}\label{eq:stage-split}
 n_1=\lfloor\rho_1n\rfloor,
 \qquad
 n_2=\lfloor\rho_2n\rfloor,
 \qquad
 n_3=n-n_1-n_2,
\end{equation}
where $\rho_s>0$ and $\rho_1+\rho_2+\rho_3=1$.  Our default is
$(\rho_1,\rho_2,\rho_3)=(0.35,0.30,0.35)$.
Each person contributes to exactly one stage; the final score vectors are fresh relative to both learning stages.

Initialize deterministic $q_{0,j}\in[L,U]$.  Choose $a\in(1/2,1)$ and $\gamma_t=\gamma_0t^{-a}$ with $\gamma_0>0$.  At Stage-1 arrival $t$, the server sorts the current thresholds $q_{t-1,1},\ldots,q_{t-1,k_n}$, retains their target labels, and sends the ordered list to the user.  The user releases one tree-OUE report.  If target $j$ has current rank $r_t(j)$, the server decodes using \eqref{eq:fused-score}
\[
 G_{t,j}^{(1)}=G_{r_t(j)}(\tau_j)
\]

after which it updates
\begin{equation}\label{eq:sgd-update}
 q_{t,j}=\Pi_{[L,U]}\left[q_{t-1,j}-\gamma_t\{G_{t,j}^{(1)}-\tau_j\}\right].
\end{equation}
Here $\Pi_{[L,U]}$ clips an iterate to the public bracket.  The score is conditionally unbiased:
\[
 \E\{G_{t,j}^{(1)}\mid\cF_{t-1}\}=F(q_{t-1,j}),
\]
where $\cF_{t-1}$ contains the previous private transcript.  Stable tags keep each target attached to its CDF score through coordinate crossings.

The preliminary center is the full Polyak--Ruppert average
\begin{equation}\label{eq:full-average}
 q_j^0=\ol q_{n_1,j}=\frac1{n_1}\sum_{t=1}^{n_1}q_{t,j}.
\end{equation}
Both the iterates and their average are updated in $O(k_n)$ memory.  All iterates enter the average; the online identity and tail-average comparison are given in \Cref{sec:streaming}.

\subsection{Density learning and fresh-score calibration}\label{sec:stage23}

Stage 2 freezes the preliminary centers and queries $q_j^0\pm h_n$ with deterministic bandwidth $h_n>0$.  The $2k_n$ tagged thresholds are sorted once.  Every Stage-2 user sends one tree-OUE report, from which the server obtains fused scores $G_{i,j,-}^{(2)}$ and $G_{i,j,+}^{(2)}$.  Let
\[
 \ol G_{j,\pm}^{(2)}=\frac1{n_2}\sum_{i=1}^{n_2}G_{i,j,\pm}^{(2)}
\]
and define
\begin{equation}\label{eq:density-raw}
 \wt f_j=\frac{\ol G_{j,+}^{(2)}-\ol G_{j,-}^{(2)}}{2h_n}.
\end{equation}
For a deterministic sequence $\varpi_n\downarrow0$, use the vanishing stabilization
\begin{equation}\label{eq:density-clip}
 \wh f_j=\min\{\varpi_n^{-1},\max(\varpi_n,\wt f_j)\}.
\end{equation}
Consistency makes the clipping inactive with probability tending to one.

Stage 3 again freezes the tagged list $q_1^0,\ldots,q_{k_n}^0$.  From each new user's single report, the server obtains a vector
\[
 G_i^{(3)}=(G_{i,1}^{(3)},\ldots,G_{i,k_n}^{(3)})^\top,
 \qquad
 \ol G_j^{(3)}=n_3^{-1}\sum_{i=1}^{n_3}G_{i,j}^{(3)}.
\]
The one-step estimator is
\begin{equation}\label{eq:onestep}
 \check Q_j=q_j^0+\frac{\tau_j-\ol G_j^{(3)}}{\wh f_j}.
\end{equation}
Conditional on Stages 1 and 2, the Stage-3 vectors are independent and identically distributed.  Write
\[
 \wh\sigma_{G,j}^2
 =\frac1{n_3}\sum_{i=1}^{n_3}
   (G_{i,j}^{(3)}-\ol G_j^{(3)})^2.
\]
Write $\widehat{\mathcal J}_n=\{j:\wh\sigma_{G,j}>0\}$; a zero-variance coordinate receives $[L,U]$.  Take an empty maximum as zero.
For each draw $b$, one independent $\xi_i^{(b)}\sim N(0,1)$
multiplies all coordinates of user $i$'s centered score vector.  Sharing
this multiplier across coordinates preserves their empirical correlation
in the simulated vector.  For $1\le b\le M$, define
\begin{equation}\label{eq:bootstrap-stat}
 T_b^*=\max_{j\in\widehat{\mathcal J}_n}
 \left|
 \frac{n_3^{-1/2}\sum_{i=1}^{n_3}
 \xi_i^{(b)}(G_{i,j}^{(3)}-\ol G_j^{(3)})}
 {\wh\sigma_{G,j}}
 \right|.
\end{equation}
Let $T_{(1)}^*\le\cdots\le T_{(M)}^*$ be the ordered multiplier statistics and set
\begin{equation}\label{eq:bootstrap-quantile}
 r_{M,\alpha}=\min\{M,\lceil(M+1)(1-\alpha)\rceil\},
 \qquad
 \wh c_{1-\alpha}^*=T_{(r_{M,\alpha})}^*.
\end{equation}
The simultaneous intervals are
\begin{equation}\label{eq:intervals}
 \cI_{n,j}=
 \left[
 \check Q_j-\frac{\wh c_{1-\alpha}^*\wh\sigma_{G,j}}
 {\wh f_j\sqrt{n_3}},
 \check Q_j+\frac{\wh c_{1-\alpha}^*\wh\sigma_{G,j}}
 {\wh f_j\sqrt{n_3}}
 \right]\cap[L,U].
\end{equation}

The same reporting rule serves all three stages, with $k_n$ thresholds for localization and inference and $2k_n$ for density estimation. Its unary message uses $O(k_n)$ bits. Maintaining running score sums and multiplier-weighted sums gives $O(Mk_n)$ server memory, independent of the horizon; \Cref{sec:streaming} gives the identities.

For implementation, the default split is $(0.35,0.30,0.35)$, Stage 1 uses its full average, and the gain is $0.1(U-L)t^{-2/3}$. Initial values linearly interpolate the target probabilities across the public bracket. The bandwidth is the mechanism-specific expression in \Cref{sec:simulation} with multiplier one, and stabilization is $\varpi_n=\{\log(n+e)\}^{-3/2}$. At $M=500$ and $\alpha=0.05$, \eqref{eq:bootstrap-quantile} uses rank 476 and the fixed-$M$ coverage limit is $476/501\approx0.9501$.

For the normal design, write $\Phi$ and $\phi$ for the standard normal CDF and density. For both standard-normal designs, $L_F=\phi(0)=1/\sqrt{2\pi}$ is admissible, so the main-design gain $\gamma_0\approx0.411$ and the CDF-comparison gain $\gamma_0=0.6$ satisfy the gain condition in \Cref{ass:gain}.

\begin{algorithm}[tbp]
\singlespacing
\caption{Fused one-pass LDP inference for a growing quantile grid}\label{alg:main}
\begin{algorithmic}[1]
\Require horizon $n$; target probabilities $\tau_1<\cdots<\tau_{k_n}$; public bracket $[L,U]$; privacy level $\eps_n$; split $(\rho_1,\rho_2,\rho_3)$; gain $(\gamma_0,a)$; bandwidth $h_n$; stabilization $\varpi_n$; level $\alpha$; multiplier count $M\ge1$.
\Statex Process the cohorts consecutively, indexing users from one within each cohort. The public arrival index alone determines an incoming user's stage.
\State Compute $(n_1,n_2,n_3)$ from \eqref{eq:stage-split}. Initialize deterministic $q_{0,j}\in[L,U]$ and $\ol q_{0,j}=0$.
\State \textbf{For arrivals $i=1,\ldots,n_1$ (Stage 1):} sort the current tagged thresholds, obtain one $\eps_n$-LDP tree-OUE report, decode all fused scores with anchors $\tau_j$, update \eqref{eq:sgd-update}, and update the full average by \eqref{eq:recursive-average}. Set $q_j^0=\ol q_{n_1,j}$.
\State \textbf{For arrivals $i=1,\ldots,n_2$ (Stage 2):} use the fixed tagged list $q_j^0\pm h_n$, obtain one report per user, and update the $2k_n$ running score sums. Compute $\wt f_j$ and $\wh f_j$ from \eqref{eq:density-raw}--\eqref{eq:density-clip}.
\State Before Stage 3, independently generate multiplier streams $\{\xi_i^{(b)}:1\le i\le n_3,1\le b\le M\}$ from a public random seed.
\State \textbf{For arrivals $i=1,\ldots,n_3$ (Stage 3):} use the fixed list $q_j^0$, obtain one report, and update for every $j$ the running sums $\sum_iG_{i,j}^{(3)}$, $\sum_i(G_{i,j}^{(3)})^2$, and $\sum_i\xi_i^{(b)}G_{i,j}^{(3)}$ for $b\le M$, together with $\sum_i\xi_i^{(b)}$.
\State Compute $\check Q_j$, $\wh\sigma_{G,j}$, $T_b^*$, and the intervals in \eqref{eq:onestep}--\eqref{eq:intervals}. Take the maximum over $\widehat{\mathcal J}_n=\{j:\wh\sigma_{G,j}>0\}$, using zero if this set is empty; return $[L,U]$ for excluded coordinates. Optionally apply the monotone interval refinement in \Cref{prop:monotone-refinement}.
\Ensure $\{\check Q_j,\cI_{n,j}:1\le j\le k_n\}$.
\end{algorithmic}
\end{algorithm}

The algorithm is online in the one-pass sense and uses memory independent of the stream length for fixed $(M,k_n)$.  Stage 1 adapts to past reports, whereas Stages 2 and 3 use frozen thresholds.  The three boundaries and the coverage statement refer to a horizon fixed before data collection.  Scalar confidence-sequence constructions use dynamically growing parallel paths for asymptotic time-uniform validity \citep{Cai2025}; the present fixed-terminal construction adds simultaneous calibration over a growing quantile grid.

\subsection{Streaming memory and monotone post-processing}\label{sec:streaming}\label{app:implementation}

For each Stage-3 coordinate, a memory-bounded implementation maintains
\[
 S_{1,j}=\sum_{i=1}^{n_3}G_{i,j}^{(3)},
 \qquad
 S_{2,j}=\sum_{i=1}^{n_3}(G_{i,j}^{(3)})^2.
\]
Then
\[
 \ol G_j^{(3)}=S_{1,j}/n_3,
 \qquad
 \wh\sigma_{G,j}^2=S_{2,j}/n_3-(S_{1,j}/n_3)^2.
\]
For multiplier draw $b$, maintain
\[
 R_{b,j}=\sum_{i=1}^{n_3}\xi_i^{(b)}G_{i,j}^{(3)},
 \qquad
 r_b=\sum_{i=1}^{n_3}\xi_i^{(b)}.
\]
The centered numerator in \eqref{eq:bootstrap-stat} is
\[
 n_3^{-1/2}(R_{b,j}-r_bS_{1,j}/n_3).
\]
These identities give $O(Mk_n)$ server memory, independent of the stream length. Vanilla tree-OUE uses $O(k_n)$ report bits. Compact alternatives such as optimized local hashing or Hadamard response require a channel-specific fusion covariance \citep{Wang2017,Acharya2019}.  Floating-point
implementations can compute variances with Welford's stable online
recurrence.  The simulation code returns $[L,U]$ for all coordinates if any estimated score variance is numerically zero or the covariance factorization fails.

Because $Q_1\le\cdots\le Q_{k_n}$, define from raw intervals $[L_j^I,U_j^I]$
\[
 \wt L_j^I=\max_{\ell\le j}L_\ell^I,
 \qquad
 \wt U_j^I=\min_{\ell\ge j}U_\ell^I.
\]

\begin{proposition}[Coverage-preserving monotone refinement]\label{prop:monotone-refinement}
If $Q_j\in[L_j^I,U_j^I]$ for every $j$, then
$Q_j\in[\wt L_j^I,\wt U_j^I]$ for every $j$.
\end{proposition}

The full Stage-1 average is maintained without storing the path:
\begin{equation}\label{eq:recursive-average}
 \ol q_{t,j}=\ol q_{t-1,j}+t^{-1}(q_{t,j}-\ol q_{t-1,j}).
\end{equation}
The full average uses every projected iterate and preserves the entire Stage-1 sample.

\begin{proposition}[Full versus delayed averaging]\label{prop:full-tail}
For $0\le b<n_1$, let
\[
 \ol q_{n_1,j}^{(b)}
 =(n_1-b)^{-1}\sum_{t=b+1}^{n_1}q_{t,j}.
\]
Then, deterministically,
\[
 \max_{j\le k_n}|\ol q_{n_1,j}^{(b)}-\ol q_{n_1,j}|
 \le (U-L)\frac{b}{n_1}.
\]
\end{proposition}

Thus full and delayed averages are equivalent at rate $r_{1n}$ when $b/n_1=o(r_{1n})$.

The active set $\widehat{\mathcal J}_n$ and the zero-variance fallback are defined in \Cref{sec:stage23}.  By
\Cref{lem:sample-scale,lem:variance-bounds},
\[
 \Pp\{\widehat{\mathcal J}_n=\{1,\ldots,k_n\}\}\longrightarrow1.
\]

\section{Theoretical guarantees}\label{sec:theory}

We state the regularity and growth conditions, then give the learning rates and simultaneous inference guarantees. 

\subsection{Assumptions}\label{sec:assumptions}

Write $\ell_n=\log\{2k_n(n+1)\}$.  For a threshold list of length $d$, use $H_d$, $\nu_{\eps_n}$, and $\cB_{d,\eps_n}$ from \eqref{eq:nu-envelope}.  Set $
 B_{1n}=\cB_{k_n,\eps_n},
 B_{2n}=\cB_{2k_n,\eps_n}.$

\begin{assumption}[Sampling, targets, and interiority]\label{ass:iid}
The observations are i.i.d. with continuous CDF $F$.  There are constants
$\delta_0>0$ and $0<\tau_-<\tau_+<1$ such that, for every $n$,
\[
 \tau_-\le \tau_{1,n}<\cdots<\tau_{k_n,n}\le\tau_+,
 \qquad
 Q_{j,n}\in[L+3\delta_0,U-3\delta_0]
 \quad (1\le j\le k_n).
\]
Every user participates exactly once and uses independent local randomization conditional on the public transcript.
\end{assumption}

The target spacing may decrease with $n$. Stable tags preserve target
identities when close thresholds cross during Stage 1 or coincide after
finite-precision rounding, and the score remains unbiased at each tagged
threshold. The empirical multiplier incorporates the resulting
cross-quantile correlation.

\begin{assumption}[Local quantile regularity]\label{ass:smooth}
For some $\beta\in(0,1]$, the CDF is absolutely continuous on
\[
 \cN_n=\bigcup_{j\le k_n}[Q_{j,n}-2\delta_0,Q_{j,n}+2\delta_0]
\]
with density $f$ satisfying, uniformly on $\cN_n$,
\[
 0<f_-\le f(x)\le f_+<\infty,
 \qquad
 |f(x)-f(y)|\le L_f|x-y|^\beta.
\]
The constants do not depend on $n$.
\end{assumption}

Positivity and local continuity of the density are standard regular-quantile
conditions for Bahadur and one-step expansions
\citep{Bahadur1966,VanderVaart1998}; closely related local conditions are
used for online locally private quantile inference in \citet{Liu2023}.
The lower bound $f_->0$ identifies each quantile locally and converts
CDF error into quantile error. The $\beta$-H\"older bound controls both
the density bias in \eqref{eq:density-bias} and the one-step remainder
in \eqref{eq:onestep-two-remainders}, uniformly over the growing grid.
A locally Lipschitz density gives $\beta=1$; a common modulus of
continuity can be used in the same arguments when explicit polynomial
rates are less important. Extreme targets with vanishing density require
the corresponding nonregular quantile normalization.

To control Stage 1 from any initialization in the projection bracket, we use the following secant condition.

\begin{assumption}[Uniform Stage-1 identification]\label{ass:identification}
There are constants $0<\mu\le L_F<\infty$ such that for every $n$, every $j\le k_n$, and every $q\in[L,U]$,
\begin{equation}\label{eq:secant}
 \mu|q-Q_{j,n}|
 \le |F(q)-\tau_{j,n}|
 \le L_F|q-Q_{j,n}|,
 \qquad
 (q-Q_{j,n})\{F(q)-\tau_{j,n}\}\ge0.
\end{equation}
\end{assumption}

\Cref{ass:identification} is the monotone root-identification condition for
the projected Robbins--Monro step on $[L,U]$.  Together with the decreasing
gain in \Cref{ass:gain}, it follows the standard stochastic-approximation
setup of \citet{RobbinsMonro1951,PolyakJuditsky1992,KushnerYin2003}.
One sufficient condition is absolute continuity on $[L,U]$ with
$0<\mu\le f(x)\le L_F<\infty$ almost everywhere, since
$F(q)-F(Q_j)=\int_{Q_j}^q f(x)\,dx$ gives \eqref{eq:secant}.
In fact, local regularity in \Cref{ass:smooth} suffices on the fixed
bracket. For $|q-Q_j|\le\delta_0$, integration gives secant bounds
$f_-$ and $f_+$. If $q\ge Q_j+\delta_0$, monotonicity gives
\[
 \frac{f_-\delta_0}{U-L}
 \le \frac{F(q)-F(Q_j)}{q-Q_j}
 \le \frac1{\delta_0},
\]
and the same bounds hold on the left of $Q_j$. Thus \eqref{eq:secant}
holds with $\mu=\min\{f_-,f_-\delta_0/(U-L)\}$ and
$L_F=\max\{f_+,1/\delta_0\}$. The burn-in conditions below then
control early private-noise excursions.

The index $b_n$ below marks entry into the local neighborhood in the analysis; the implemented full average retains every Stage-1 iterate.
\begin{assumption}[Gain, stage sizes, and localization]\label{ass:gain}
The gain is $\gamma_t=\gamma_0t^{-a}$ with $a\in(1/2,1)$ and fixed
\[
 0<\gamma_0\le \frac{1}{2L_F}.
\]
The initial values are deterministic in $[L,U]$.  There is a deterministic integer $b_n\in\{1,\ldots,n_1-1\}$ satisfying
\begin{align}
 &b_n/n_1\to0,\label{eq:bn1}\\
 &\exp\left(-\frac{\mu}{2}\sum_{t=1}^{b_n}\gamma_t\right)(U-L)
 =o\{\min(r_{1n},\delta_0)\},\label{eq:bn2}\\
 &B_{1n}\left\{\sqrt{\gamma_{b_n}\ell_n}+\gamma_{b_n}\ell_n\right\}=o(\delta_0),\label{eq:bn3}
\end{align}
Here $r_{1n}$ is defined in \eqref{eq:r1n} below.  Together with
\Cref{ass:iid,ass:identification}, conditions
\eqref{eq:bn1}--\eqref{eq:bn3} give uniform entry into the local
neighborhood and eventual inactivity of the projection; a direct choice of
$b_n$ is given in \Cref{lem:localization}.
\end{assumption}

The explicit preliminary rate is
\begin{equation}\label{eq:r1n}
\begin{split}
 r_{1n}={}&\frac{b_n}{n_1}
 +B_{1n}\sqrt{\frac{\ell_n}{n_1}}
 +B_{1n}n_1^{a/2-1}\sqrt{\ell_n}\\
 &+B_{1n}^{1+\beta}n_1^{-a(1+\beta)/2}
   \ell_n^{(1+\beta)/2}.
\end{split}
\end{equation}
The four terms correspond to the initial segment, the averaged martingale,
the Abel endpoint, and the nonlinear H\"older remainder.

\begin{assumption}[Density stage]\label{ass:density}
The bandwidth satisfies $h_n\to0$, $h_n<\delta_0$, and
\[
 d_n=r_{1n}^\beta+h_n^\beta
 +\frac{B_{2n}}{h_n}
 \left(\sqrt{\frac{\ell_n}{n_2}}+\frac{\ell_n}{n_2}\right)
 \longrightarrow0.
\]
The deterministic clipping sequence obeys $\varpi_n\downarrow0$ and $\varpi_n^{-1}\to\infty$.
\end{assumption}

Let $p_j^0=F(q_j^0)$ and, for a Stage-3 observation, define
\[
 Y_{i,j}=G_{i,j}^{(3)}-p_j^0,
 \qquad
 \sigma_{j,n}^2=\E(Y_{i,j}^2\mid\cF_{n_1+n_2}).
\]
Let
\[
 \ul\sigma_n=\min_{j\le k_n}\sigma_{j,n},
 \qquad
 K_n=\frac{2B_{1n}}{\ul\sigma_n}.
\]
By \Cref{prop:fused-moments}, $|Y_{i,j}|/\sigma_{j,n}\le K_n$
almost surely conditional on the first two stages.  The resulting support
and variance bounds are established in \Cref{lem:variance-bounds}.

\begin{assumption}[High-dimensional approximation and one-step remainder]\label{ass:bootstrap}
The following random rates converge to zero in probability:
\begin{align}
 &\left(\frac{K_n^2\ell_n^7}{n_3}\right)^{1/6}
 =o_{\Pp}(1),
 \label{eq:hdclt-cond}\\
 &\frac{\sqrt{n_3\ell_n}}{\ul\sigma_n}
 \{r_{1n}d_n+r_{1n}^{1+\beta}\}
 =o_{\Pp}(1).
 \label{eq:remainder-cond}
\end{align}
For an empirical multiplier quantile based on $M_n$ draws, assume $M_n\to\infty$.  The exact conditional multiplier quantile is the theoretical reference.
\end{assumption}

The first condition is the Gaussian approximation condition for bounded
high-dimensional vectors in \citet{CCK2017}; the second makes the one-step
remainder negligible on the maximum scale.  \Cref{cor:width} gives a simple
growth regime that implies both conditions.

\subsection{Localization and density rates}

The Stage-1 and Stage-2 arguments are collected in
\Cref{app:stage1,app:onestep}.  They establish the uniform rates
$\max_j|q_j^0-Q_{j,n}|=O_{\Pp}(r_{1n})$ and
$\max_j|\wh f_j-f(Q_{j,n})|=O_{\Pp}(d_n)$; see
\Cref{thm:preliminary,thm:density}.  The term $r_{1n}^{\beta}$ in $d_n$
accounts for centering the finite difference at the estimated quantile.
For $\beta=1$, balancing its bandwidth and stochastic terms gives
$h_n\asymp\{B_{2n}^2\ell_n/n_2\}^{1/4}$.

\begin{theorem}[Full-average preliminary rate]\label{thm:preliminary}
Suppose \Cref{ass:iid,ass:smooth,ass:identification,ass:gain} hold.  Then
the full online average in \eqref{eq:full-average} satisfies
\begin{equation}\label{eq:prelim-rate}
 \max_{j\le k_n}|q_j^0-Q_{j,n}|=O_{\Pp}(r_{1n}).
\end{equation}
On an event with probability tending to one, it also admits the expansion
\begin{equation}\label{eq:stage1-expansion}
 f(Q_{j,n})(q_j^0-Q_{j,n})
 =-\frac1{n_1}\sum_{t=b_n+1}^{n_1}\zeta_{t,j}
   +R_{1n,j},
\end{equation}
where $\zeta_{t,j}=G_{t,j}^{(1)}-F(q_{t-1,j})$ and
$\max_j|R_{1n,j}|$ has the order given by the early-iterate,
Abel-endpoint, and nonlinear terms in \eqref{eq:r1n}.
\end{theorem}

\begin{theorem}[Uniform private density rate]\label{thm:density}
Suppose \Cref{ass:iid,ass:smooth,ass:density} hold and
\eqref{eq:prelim-rate} is valid.  Then
\begin{equation}\label{eq:density-rate}
 \max_{j\le k_n}|\wh f_j-f(Q_{j,n})|=O_{\Pp}(d_n).
\end{equation}
With probability tending to one, all $q_j^0\pm h_n$ lie in the interior
neighborhood and both clipping operations in \eqref{eq:density-clip} are
inactive.
\end{theorem}

In applications, the minimum estimated density, the number of active
clipping operations, and bandwidth sensitivity help identify quantiles
with weak local information.

\subsection{One-step expansion and simultaneous coverage}

\begin{theorem}[Uniform one-step representation]\label{thm:onestep}
Suppose \Cref{ass:iid,ass:smooth,ass:density} hold and \eqref{eq:prelim-rate} and \eqref{eq:density-rate} are valid.  Then
\begin{equation}\label{eq:onestep-representation}
 \max_{j\le k_n}
 \left|
 \wh f_j(\check Q_j-Q_{j,n})+
\frac1{n_3}\sum_{i=1}^{n_3}Y_{i,j}
 \right|
 =O_{\Pp}(R_n),
 \qquad
 R_n=r_{1n}d_n+r_{1n}^{1+\beta}.
\end{equation}
\end{theorem}

The conditional mean of $\tau_j-\ol G_j^{(3)}$ is $\tau_j-F(q_j^0)$. Dividing by $\wh f_j$ therefore cancels the linear location error. The remainder $R_n$ has a direct interpretation: estimating the derivative costs $r_{1n}d_n$, and linearizing $F$ costs $r_{1n}^{1+\beta}$. The simultaneous transfer condition above makes these errors small even after taking the maximum over targets.

Let $c_{1-\alpha}^*$ denote the exact conditional $(1-\alpha)$ quantile of \eqref{eq:bootstrap-stat} given the observed Stage-3 scores and the first two stages.  Let $\cI_{n,j}^*$ denote the interval in \eqref{eq:intervals} with $\wh c_{1-\alpha}^*$ replaced by $c_{1-\alpha}^*$.

\begin{theorem}[Growing-grid multiplier validity]\label{thm:bootstrap}
Suppose \Cref{ass:iid,ass:smooth,ass:identification,ass:gain,ass:density,ass:bootstrap} hold.  Define
\[
 T_n=\max_{j\in\widehat{\mathcal J}_n}
 \left|
 \frac{\sqrt{n_3}\wh f_j(\check Q_j-Q_{j,n})}
 {\wh\sigma_{G,j}}
 \right|.
\]
Then, for every fixed $0<\alpha_0<1/2$,
\begin{equation}\label{eq:bootstrap-validity}
 \sup_{\alpha\in[\alpha_0,1-\alpha_0]}
 \left|
 \Pp\{T_n\le c_{1-\alpha}^*\}-(1-\alpha)
 \right|\longrightarrow0.
\end{equation}
Consequently,
\begin{equation}\label{eq:coverage}
 \Pp\{Q_{j,n}\in\cI_{n,j}^*\text{ for all }j\le k_n\}
 \longrightarrow1-\alpha.
\end{equation}
If the empirical quantile based on $M_n\to\infty$ multiplier draws is used, then \eqref{eq:bootstrap-validity} and \eqref{eq:coverage} remain valid with $\wh c_{1-\alpha}^*$ and $\cI_{n,j}$ in place of $c_{1-\alpha}^*$ and $\cI_{n,j}^*$, respectively.  Here the probability also averages over the independent calibration multipliers.
If instead $M$ is fixed, then for every fixed $\alpha\in(0,1)$,
\begin{equation}\label{eq:fixed-M-coverage}
 \Pp\{T_n\le\wh c_{1-\alpha}^*\}
 \longrightarrow \frac{r_{M,\alpha}}{M+1},
\end{equation}
and the simultaneous coverage probability has the same limit.
\end{theorem}

The multiplier uses the dependence among neighboring quantiles through each user's score vector, calibrating their maximum error jointly.

The fixed positive cohort fractions used by the procedure give a simple
allocation satisfying the stage-size requirements. More generally, the
proofs require $n_1,n_2,n_3\to\infty$ together with
\Cref{ass:gain,ass:density,ass:bootstrap}. In particular, $n_2/n\to0$
is allowed when the density consistency and one-step remainder conditions
remain valid, including
\[
 \frac{B_{2n}}{h_n}\sqrt{\frac{\ell_n}{n_2}}\to0.
\]
Increasing $n_3$ improves first-order precision after Stage 1 has localized
the targets and Stage 2 has estimated their densities accurately. This is
the tradeoff illustrated by the allocation results in \Cref{sec:sim-e2e}:
smaller learning cohorts can increase the remainder even while $n_3$
reduces the leading variance.

\begin{corollary}[Width and grid growth]\label{cor:width}
Under the assumptions of \Cref{thm:bootstrap},
\begin{equation}\label{eq:width-general}
 \max_{j\le k_n}\operatorname{halfwidth}(\cI_{n,j})
 =O_{\Pp}\left
 \{\frac{H_{k_n}\sqrt{1+\nu_{\eps_n}}\sqrt{\log(2k_n)}}
 {f_-\sqrt{n_3}}\right\}.
\end{equation}
For $0<\eps_n\le1$, this is
\begin{equation}\label{eq:width-small-eps}
 O_{\Pp}\left\{
 \frac{H_{k_n}\sqrt{\log(2k_n)}}
 {\eps_nf_-\sqrt{n_3}}
 \right\}.
\end{equation}
If $\eps_n$ is bounded away from zero, $n_s/n\to\rho_s\in(0,1)$, $a=2/3$, $\beta=1$, and
\begin{equation}\label{eq:transparent-growth}
 \frac{\log^{9}\{2k_n(n+1)\}}{n}\longrightarrow0,
\end{equation}
then the conditions of \Cref{thm:bootstrap} hold for the bandwidth
\[
 h_n\asymp\left\{\frac{B_{2n}^2\ell_n}{n_2}\right\}^{1/4}
\]
provided $b_n/n=o(\ell_n^{3/2}n^{-1/2})$.  In particular, every polynomially growing $k_n$ is allowed.
\end{corollary}

The width bound in \eqref{eq:width-general} separates the
private-score scale $H_{k_n}\sqrt{1+\nu_{\eps_n}}$ from the simultaneous
calibration factor $\sqrt{\log(2k_n)}$.  The density lower bound converts
CDF error into quantile error.  Thus, at fixed privacy and density bounds,
grid size enters this upper bound only through logarithmic factors, while
precision improves at the usual square-root rate in the inference cohort
size $n_3$.
The general bound also allows $\eps_n$ to vary through the score
variance and envelope. For small $\eps_n$,
$\nu_{\eps_n}\asymp\eps_n^{-2}$ and
$\cB_{d,\eps_n}\asymp H_d\eps_n^{-1}$, displaying the effective
private sample size in each rate.
Condition \eqref{eq:transparent-growth} combines the tree-score envelope,
density estimation, Gaussian approximation, and anti-concentration in one
simple rate.  It allows
$k_n=\exp\{o(n^{1/9})\}$, including every polynomially growing grid.

\section{Fixed-chain calibration}\label{app:fixed-results}

\subsection{Restarted-chain procedures}\label{app:legacy}

This subsection defines the blockwise procedures used in the calibration
comparison.  Let $J$ be the number of blocks and
$\wh Q_b=(\wh Q_{b,1},\ldots,\wh Q_{b,k_n})^\top$ the averaged
private stochastic-approximation output from block $b$.  The block
aggregate and coordinate scales are
\[
 \wt Q=J^{-1}\sum_{b=1}^J\wh Q_b,
 \qquad
 S_j^2=J^{-1}\sum_{b=1}^J(\wh Q_{b,j}-\wt Q_j)^2.
\]
The raw chain multiplier statistic is
\[
 T_{\mathrm{chain}}^*
 =\max_{j\le k_n}
 \left|
 \frac{J^{-1/2}\sum_{b=1}^J\xi_b
 (\wh Q_{b,j}-\wt Q_j)}{S_j}
 \right|.
\]
This raw multiplier approximates the large-$J$ regime in which block bias
is small relative to $S_j/\sqrt J$.

Under a multivariate normal block model, the sample mean is independent of
the Wishart sample covariance.  The fixed-$J$ Student--Wishart comparator
simulates this pivot using the public balanced tree-OUE score correlation,
thereby retaining the random coordinate scales and cross-coordinate
dependence.  For restarted stochastic-approximation outputs, it gives the
corresponding first-order Gaussian calibration.

More explicitly, let $k=k_n$, let $R$ be the public score correlation matrix, and put $\nu=J-1$.  Let
$V\sim\operatorname{Wishart}_k(R,\nu)$ be independent of
$Z\sim N_k(0,R)$.  Orthogonally separating the block mean from the
$\nu$ residual contrasts gives the exact Gaussian-model maximum
\[
 \max_{j\le k}\frac{|Z_j|}{\sqrt{V_{jj}/\nu}}.
\]
This representation remains valid for singular $R$.  Its critical value
depends on $R$: if $R=\boldsymbol 1\boldsymbol 1^\top$, all coordinates
coincide and the maximum has the $|t_\nu|$ distribution for every $k$.
The polynomial rate in \Cref{prop:bonferroni} therefore applies to the
Student--Bonferroni calibration; the Student--Wishart rate is determined by
the correlation matrix $R$.

The marginal Student--Bonferroni safeguard uses
\[
 \wt Q_j\pm t_{J-1,1-\alpha/(2k_n)}\frac{s_j}{\sqrt J},
 \qquad
 s_j^2=(J-1)^{-1}\sum_{b=1}^J(\wh Q_{b,j}-\wt Q_j)^2,
\]
Together with the raw multiplier, these intervals distinguish fixed-$J$
tail calibration from blockwise
estimation bias.  Their polynomial critical-value growth is given in
\Cref{prop:bonferroni}.

The common block bias and fixed-$J$ target distribution remain unchanged as the number of multiplier draws increases. Increasing the number of chains with the horizon yields the distinct time-uniform construction of \citet{Cai2025}.

\subsection{Calibration with a fixed number of restarted chains}\label{sec:fixedJ}

\begin{proposition}[Fixed-$J$ Gaussian-multiplier mismatch]\label{prop:fixedJ}
Let $W_1,\ldots,W_J$ be i.i.d. $N(\theta,\sigma^2)$ with fixed $J\ge2$, and let
\[
 \ol W=J^{-1}\sum_{b=1}^JW_b,
 \qquad
 S_J^2=J^{-1}\sum_{b=1}^J(W_b-\ol W)^2.
\]
Then
\[
 \frac{\sqrt{J-1}(\ol W-\theta)}{S_J}\sim t_{J-1}.
\]
In contrast, conditional on the observations,
\[
 \frac{J^{-1/2}\sum_{b=1}^J\xi_b(W_b-\ol W)}{S_J}
 \sim N(0,1).
\]
The discrepancy persists as the within-block sample size grows whenever $J$ remains fixed.
\end{proposition}

\begin{proposition}[Polynomial Student--Bonferroni growth]\label{prop:bonferroni}
For fixed degrees of freedom $\nu$ and fixed $\alpha\in(0,1)$,
\begin{equation}\label{eq:t-growth}
 t_{\nu,1-\alpha/(2k)}\asymp k^{1/\nu}
 \qquad (k\to\infty).
\end{equation}
Thus a fixed-$J$ Student--Bonferroni band has a critical value of
polynomial order $k^{1/(J-1)}$, and Holm's \citeyearpar{Holm1979} most stringent step has the same
fixed-degree tail.  A Gaussian maximum over $k$ standardized coordinates
has critical value at most of order $\sqrt{\log k}$ under arbitrary
correlation.
\end{proposition}

The two propositions separate calibration from estimation.
Student--Bonferroni has the polynomial critical-value growth in
\eqref{eq:t-growth}.  Student--Wishart instead uses the public first-order
score correlation induced by balanced tree OUE, so its critical value
depends on that correlation matrix.  Both corrected fixed-$J$ procedures
produce wider intervals in the numerical comparisons, whereas the
user-level multiplier uses $n_3$ independent score vectors.

\subsection{Connections with stochastic-approximation inference}\label{app:connections}

\begingroup
The construction also relates to uncertainty quantification for stochastic approximation.  HiGrad creates hierarchically split trajectories and $t$-based contrasts \citep{SuZhu2023}, while \citet{ZhuLouWeiWu2026} analyze high-confidence $t$ inference from a small number of independent stochastic-optimization runs.  \citet{DetteGraw2024} develop block-bootstrap inference for stochastic gradients under pure local differential privacy.  \citet{Xie2025} use locally perturbed gradients with Gaussian differential privacy and derive a functional central limit theorem.  Growing-grid inference additionally requires an approximation for the maximum over quantile coordinates.  We obtain it by using the first path for localization and an independent cohort of user-level scores for inference.  The private channel builds on randomized response, frequency estimation, and range-query mechanisms \citep{Warner1965,Kasiviswanathan2011,Kairouz2014,Wang2017,Duchi2018,Acharya2019,Cormode2019}; the localization proof uses classical stochastic approximation and averaging \citep{RobbinsMonro1951,Ruppert1988,PolyakJuditsky1992,KushnerYin2003}; and the final calibration invokes high-dimensional Gaussian and multiplier approximations \citep{CCK2013,CCK2017}.
\endgroup

\section{Numerical experiments}\label{sec:simulation}

We distinguish precision gained by decoding the same private report from the performance of complete inferential procedures. Write $k=k_n$. Each setting uses 500 independent Monte Carlo repetitions, 500 Gaussian multiplier draws, and nominal simultaneous coverage $0.95$. Probabilities are equally spaced on $[0.05,0.95]$. Coverage records whether the entire grid is covered in a repetition; at 95\%, its Monte Carlo standard error is about one percentage point. For point estimates $\wh Q_{r,j}$ in repetition $r$, the pooled root mean squared error is
\[
 \operatorname{RMSE}=
 \left\{(500k)^{-1}\sum_{r=1}^{500}\sum_{j=1}^k
 (\wh Q_{r,j}-Q_j)^2\right\}^{1/2}.
\]
Lower RMSE means more accurate point estimates. Average full width measures interval precision and is interpreted together with attained coverage.

\subsection{Precision from complementary decoding}\label{sec:sim-e2e}

The matched three-stage comparison uses fused tree-OUE, balanced tree-OUE, and categorical randomized response (kRR) on the interval categories, with identical cohort sizes and privacy budgets. Fused three-cohort inference is the complete procedure in \Cref{alg:main}. The two tree procedures isolate the effect of fusion; kRR replaces the reporting channel inside the same construction. Under $N(0,1)$, the public bracket is $[F^{-1}(0.02),F^{-1}(0.98)]$. Initial values interpolate linearly across that bracket, mapping the smallest and largest target probabilities to its endpoints. The defaults are the full Stage-1 average, $a=2/3$, $\gamma_0=0.1(U-L)$, split $(0.35,0.30,0.35)$, and $\varpi_n=\{\log(n+e)\}^{-3/2}$. For channel $m$ with Stage-2 envelope $B_{m,2k}$, the bandwidth is
\[
 h_{m,n}=\left\{\frac{B_{m,2k}^2\log\{2k(n+1)\}}{n_2}\right\}^{1/4},
\]
using $B_{2n}$ for tree-OUE and $(e^\eps+2k)/(e^\eps-1)$ for kRR. \Cref{sec:stage23} relates the normal-design gains to the sufficient gain condition and gives the remaining tuning details.

\begin{table}[!htbp]
\centering
\small
\caption{End-to-end sample-size scaling at $\epsilon=1$ under $N(0,1)$. Fused is the complete procedure in \Cref{alg:main}; Balanced changes its tree decoder, and kRR changes its reporting channel. All methods use the same three-cohort allocation and public bracket. Each entry is simultaneous coverage followed by pooled RMSE; lower RMSE is better.}
\label{tab:sample-size-scaling}
\begin{tabular}{rrccc}
\toprule
$k$ & $n$ & Fused & Balanced & kRR \\
\midrule
31 & 160000 & 0.932; 0.0897 & 0.926; 0.1147 & 0.918; 0.1545 \\
31 & 320000 & 0.940; 0.0602 & 0.944; 0.0766 & 0.928; 0.0949 \\
\addlinespace[2pt]
63 & 320000 & 0.944; 0.0698 & 0.938; 0.0903 & 0.892; 0.2052 \\
63 & 640000 & 0.948; 0.0490 & 0.936; 0.0625 & 0.902; 0.1340 \\
\addlinespace[2pt]
127 & 640000 & 0.948; 0.0561 & 0.952; 0.0716 & 0.844; 0.3538 \\
127 & 1280000 & 0.950; 0.0386 & 0.952; 0.0494 & 0.860; 0.2038 \\
\bottomrule
\end{tabular}
\end{table}

\Cref{tab:sample-size-scaling} separates grid growth from sample-size gains by doubling $n$ at each fixed $k$. Fusion reduces RMSE by 21--23\% relative to balanced decoding in all six settings, with coverage 93.2--95.0\%. At $k=127$ and $n=1280000$, RMSE falls from $0.0494$ to $0.0386$ at essentially the same coverage (95.2\% versus 95.0\%). This precision gain uses the same reporting channel, cohort sizes, and privacy budget. At fixed $k$, doubling $n$ reduces fused RMSE by 30--33\%, consistent with the square-root inference-sample dependence of \Cref{cor:width}. The kRR procedure has substantially larger errors and undercovers at the larger grids.

Privacy and distribution comparisons in \Cref{app:numerics} retain the fusion gain: at $(n,k)=(320000,63)$, fused RMSE decreases from $0.1490$ to $0.0326$ as $\eps$ increases from $0.5$ to $2$, with coverage 94.4--95.6\%. Under the contaminated normal design at $(640000,63,1)$, fused RMSE is $0.0842$, compared with $0.1053$ for balanced decoding. Cohort allocation matters separately: increasing the inference share to $0.55$ lowers error in regular settings but changes coverage from $0.944$ to $0.896$ at $\eps=0.5$ and from $0.938$ to $0.872$ under contamination. This allocation trades a smaller leading variance against larger learning errors.

\subsection{Privacy and distribution comparisons}\label{app:numerics}

These experiments retain the three-stage design, target probabilities, and mechanism-specific tuning in \Cref{sec:simulation}. Each cell uses 500 repetitions and 500 multipliers. The public bracket is $[F^{-1}(0.02),F^{-1}(0.98)]$ for the relevant distribution, using population-level range information. Coverage is for the complete target grid, and each table pairs it with pooled RMSE.

At fixed $(n,k)=(320000,63)$ under $N(0,1)$, varying $\eps$ isolates the privacy effect within each procedure. \Cref{tab:privacy-comparison} shows a clear precision gain from weaker privacy: fused RMSE falls from $0.1490$ to $0.0326$, with coverage 94.4--95.6\%. Fused RMSE is smaller than balanced RMSE at every privacy level.

\begin{table}[!htbp]
\centering
\small
\caption{End-to-end privacy comparison at $n=320000$ and $k=63$ under $N(0,1)$. Each method entry is simultaneous coverage followed by pooled RMSE.}
\label{tab:privacy-comparison}
\begin{tabular}{rccc}
\toprule
$\epsilon$ & Fused & Balanced & kRR \\
\midrule
0.5 & 0.944; 0.1490 & 0.932; 0.2193 & 0.874; 1.1007 \\
1.0 & 0.944; 0.0698 & 0.938; 0.0903 & 0.892; 0.2052 \\
2.0 & 0.956; 0.0326 & 0.932; 0.0402 & 0.934; 0.0474 \\
\bottomrule
\end{tabular}
\end{table}

At $(n,k,\eps)=(640000,63,1)$, we compare $N(0,1)$, a variance-standardized $t_3$ distribution, and $0.9N(0,1)+0.1N(0,4^2)$. Fused coverage is 93.8--94.8\% across the three distributions (\Cref{tab:distribution-comparison}). Contamination increases its RMSE to $0.0842$, consistent with the importance of local density for quantile precision; fusion still improves on balanced decoding, whose RMSE is $0.1053$. kRR has appreciably larger error and lower coverage under both heavy tails and contamination.

\begin{table}[!htbp]
\centering
\small
\caption{End-to-end distribution comparison at $n=640000$, $k=63$, and $\epsilon=1$. Each method entry is simultaneous coverage followed by pooled RMSE.}
\label{tab:distribution-comparison}
\begin{tabular}{lccc}
\toprule
Distribution & Fused & Balanced & kRR \\
\midrule
Normal & 0.948; 0.0490 & 0.936; 0.0625 & 0.902; 0.1340 \\
$t_3$ & 0.948; 0.0483 & 0.932; 0.0597 & 0.862; 0.1182 \\
Contamination & 0.938; 0.0842 & 0.932; 0.1053 & 0.844; 0.2377 \\
\bottomrule
\end{tabular}
\end{table}

\FloatBarrier

\subsection{Fused three-cohort inference and CDF-band inversion}\label{sec:routes}

A fixed spatial grid can use all $n$ reports to estimate CDF values and then invert a joint CDF band. We implement the private CDF route of \citet{Liu2024CDF} with our fused tree channel and multiplier calibration. The comparison uses $N(0,1)$, public bracket $[-3,3]$, $\eps=1$, and $(n,k)=(40000,31),(160000,31),(320000,63)$, with 500 repetitions and 500 multiplier draws. Both procedures use all $n$ users and one report per user. Fused three-cohort inference keeps the defaults in \Cref{sec:stage23} and learns query locations and densities from its first two cohorts. The CDF procedure instead uses $d=k$ equally spaced inspection points $x_r$ and the public anchor $u=1/2$, which preserves unbiasedness without knowing $F(x_r)$. The target grid consists of $k$ equally spaced probabilities on $[0.05,0.95]$.

Let $[\ell_r,u_r]$ be the studentized Gaussian-multiplier band for the CDF at the inspection points, clipped to $[0,1]$. We invert it using
\[
 L_j^{\mathrm{CDF}}=\max\bigl(\{L\}\cup\{x_r:u_r<\tau_j\}\bigr),\qquad
 U_j^{\mathrm{CDF}}=\min\bigl(\{U\}\cup\{x_r:\ell_r\ge\tau_j\}\bigr).
\]
On the simultaneous CDF coverage event, every resulting interval contains its target quantile. The same event gives coverage for every probability whose quantile is inside the bracket, including probabilities selected after collection. The interval endpoints retain the spatial discretization error; point estimates alone invert the piecewise-linear interpolation of the isotonic CDF estimates.

For $d$ equally spaced inspection points including $L$ and $U$, exact CDF values give a discrete inversion width of $(U-L)/(d-1)$ for every target quantile strictly inside the bracket. Indeed, if $r$ is the first index with $F(x_r)\ge\tau_j$, then $F(x_{r-1})<\tau_j$ and the endpoints are $x_{r-1}$ and $x_r$, including when $F(x_r)=\tau_j$. This separates spatial discretization from statistical uncertainty.

\begin{table}[htbp]
\centering
\small
\caption{Complete procedures under $N(0,1)$ on $[-3,3]$, with one $\eps=1$ report per user. Each entry is (simultaneous coverage; RMSE; average full width). Both methods use all $n$ users; the fused three-cohort method learns quantile locations and densities. Lower RMSE and width are preferable at comparable coverage.}\label{tab:routes}
\begin{tabular}{rrcc}
\toprule
$n$ & $k$ & Fused three-cohort & CDF inversion ($d=k$)\\
\midrule
40000 & 31 & 0.942; 0.2704; 1.1112 & 1.000; 0.1064; 0.8216\\
160000 & 31 & 0.950; 0.1134; 0.5146 & 0.994; 0.0512; 0.5067\\
320000 & 63 & 0.918; 0.0809; 0.4326 & 0.992; 0.0395; 0.3556\\
\bottomrule
\end{tabular}
\end{table}

\Cref{tab:routes} shows the resulting tradeoff. CDF inversion has smaller RMSE and narrower intervals in every setting, with conservative quantile coverage of 99.2--100\%. At $(320000,63)$, fused three-cohort coverage is 91.8\%, with a 95\% Wilson interval $[89.1,93.9]\%$ across repetitions. Coverage remains 91.8\% among runs without density clipping (\Cref{tab:direct-coverage-diagnostics}). The underlying CDF band covers its spatial grid in 93.2--97.2\% of repetitions; discrete inversion contributes the extra quantile conservatism.

For $(n,k)=(40000,31),(160000,31),(320000,63)$, direct density clipping occurs in 26.0\%, 4.8\%, and 2.0\% of repetitions; neither procedure produces fallback or empty intervals. \Cref{tab:direct-coverage-diagnostics} gives Wilson intervals for Monte Carlo uncertainty and separates repetitions with and without clipping. At $(320000,63)$, 40 of the 490 unclipped repetitions miss at least one target, so clipping alone does not explain the coverage deficit. The stratification is descriptive: clipping is an outcome of the same learning stages whose errors enter the one-step remainder.

\begin{table}[htbp]
\centering
\small
\caption{Fused three-cohort inference on $[-3,3]$ under $N(0,1)$ at $\eps=1$. All-run entries give coverage and its 95\% Wilson interval across 500 repetitions. The last two columns give covered repetitions divided by repetitions in that density-clipping group.}
\label{tab:direct-coverage-diagnostics}
\begin{tabular}{rrccc}
\toprule
$n$ & $k$ & All runs & No clipping & Clipping \\
\midrule
40000 & 31 & $0.942\ [0.918,0.959]$ & $349/370$ & $122/130$ \\
160000 & 31 & $0.950\ [0.927,0.966]$ & $452/476$ & $23/24$ \\
320000 & 63 & $0.918\ [0.891,0.939]$ & $450/490$ & $9/10$ \\
\bottomrule
\end{tabular}
\end{table}

The choice of spatial resolution also affects communication. For a report with $d$ thresholds, the privatized unary vector has expected length $(2^{H_d+1}-2)/H_d$ bits; the level tag requires another $\lceil\log_2 H_d\rceil$ bits. The CDF procedure uses $d=k$ for every user, whereas direct inference uses $d=k$ in Stages 1 and 3 and $d=2k$ in Stage 2. The 0.70/0.30 weighted mean of these two report lengths gives the direct cost under the default allocation. Both fit the same maximum report-length allowance. At $k=31$, direct inference and CDF inversion use 18.0 and 15.4 bits on average; at $k=63$, they use 28.6 and 24.0. These counts include private unary vectors and level tags, excluding public queries and transport framing.

Refining the CDF grid from $d=63$ to $127$ at $(n,k)=(320000,63)$ approximately halves its discretization width, but simulated widths remain $0.3556$ and $0.3564$, while mean report length rises to 39.3 bits. The CDF release supports later probability queries; three-cohort inference learns locations for prespecified percentiles without choosing a spatial endpoint grid. Thus the target list determines the direct method's reporting dimensions, while the public range and endpoint resolution determine the CDF grid.

Algorithm~\ref{alg:main} also updates public thresholds after each Stage-1 report; the Stage-2 and Stage-3 lists are fixed within their cohorts. Sending each user the full list requires $k(n+n_2)$ threshold values in total, or about 210 MB with 64-bit values at $(n,k)=(320000,63)$, in addition to private uploads. Counting each public list publication once gives $k(n_1+3)$ values; delivery to individual users depends on the broadcast or retrieval protocol. An equally spaced CDF grid can instead be reconstructed from its endpoints and size.

Streaming multiplier storage is $O(Md)$ for CDF inversion and $O(Mk)$ for direct inference, using the identities in \Cref{sec:streaming}. In the simulations, both implementations compute the equivalent Gaussian critical value from the empirical covariance and store $nk$ and $n_3k$ score doubles, respectively. Reported timings measure these covariance-based implementations with in-memory queries and no network transport. Upload length, query distribution, and computation therefore describe separate deployment costs.

\subsection{Restarted-block comparisons}\label{app:calibration-numerics}\label{sec:comparators}

The comparison with corrected, fixed-block inference is complementary. With $J=8$ restarted blocks, Student--Wishart and Student--Bonferroni attain 93.4--94.8\% coverage; the fused three-cohort intervals are shorter, although their point errors are larger at $k=31$ and $63$. At $k=127$, fused three-cohort and Student--Wishart widths are $0.3574$ and $0.5902$, a 39\% reduction at similar coverage and RMSE. The block design and model-assisted correlation calibration are given in \Cref{app:calibration-numerics}; the fixed-$J$ theory is in \Cref{app:fixed-results}.

\begin{table}[!htbp]
\centering
\small
\caption{Corrected block procedures versus fused three-cohort inference (\Cref{alg:main}) at $\epsilon=1$ under $N(0,1)$. Restarted methods use $J=8$ and balanced tree-OUE; the complete designs differ as specified in the text. Width is the mean full interval width.}
\label{tab:corrected-blocks}
\begin{tabular}{rrlccc}
\toprule
$n$ & $k$ & Method & Coverage & RMSE & Width \\
\midrule
\multirow{3}{*}{160000} & \multirow{3}{*}{31} & Fused three-cohort & 0.932 & 0.0897 & 0.4777 \\
 &  & Student--Wishart & 0.938 & 0.0764 & 0.6367 \\
 &  & Student--Bonferroni & 0.948 & 0.0764 & 0.6432 \\
\addlinespace[2pt]
\multirow{3}{*}{320000} & \multirow{3}{*}{63} & Fused three-cohort & 0.944 & 0.0698 & 0.4178 \\
 &  & Student--Wishart & 0.934 & 0.0662 & 0.6165 \\
 &  & Student--Bonferroni & 0.946 & 0.0662 & 0.6240 \\
\addlinespace[2pt]
\multirow{3}{*}{640000} & \multirow{3}{*}{127} & Fused three-cohort & 0.948 & 0.0561 & 0.3574 \\
 &  & Student--Wishart & 0.944 & 0.0570 & 0.5902 \\
 &  & Student--Bonferroni & 0.944 & 0.0570 & 0.5971 \\
\bottomrule
\end{tabular}
\end{table}

The block comparisons use $N(0,1)$, $\eps=1$, and $(n,k)=(160000,31)$, $(320000,63)$, $(640000,127)$, with 500 repetitions and 500 calibration draws where needed. They divide all users into $J=8$ equal blocks and use balanced tree-OUE. Each block initializes by linear interpolation across $[\Phi^{-1}(0.01),\Phi^{-1}(0.99)]$, restarts the gain $0.1(U-L)t^{-2/3}$, and averages its second half. Student--Wishart and Student--Bonferroni share the resulting point estimator; fused three-cohort inference instead uses the main three-cohort design and its $[\Phi^{-1}(0.02),\Phi^{-1}(0.98)]$ bracket. The comparison includes these decoder, bracket, path, and allocation choices.

Student--Wishart uses the public, model-based balanced tree score correlation in the Gaussian block approximation. It is a model-assisted comparator; Student--Bonferroni uses the corresponding marginal $t$ tails. \Cref{app:fixed-results} defines both references. \Cref{tab:corrected-blocks} retains the coverage--error--width tradeoff: fused three-cohort bands are shorter in all settings, while restarted estimates have smaller RMSE at $k=31$ and $63$ and similar RMSE at $k=127$.

\FloatBarrier

\section{California wage-percentile comparisons}\label{sec:application}

We illustrate wage-percentile reporting with the 2024 one-year California person file from the American Community Survey Public Use Microdata Sample (ACS PUMS) \citep{USCensus2024PUMS}.  The 90/50 and 50/10 ratios distinguish dispersion above and below the median \citep{AutorKatzKearney2006}.  Each released record receives an additional local randomization.

We retain respondents aged 25--64 with positive wages, at least 40 weeks worked, and at least 30 usual weekly hours. We adjust annual wages to 2024 dollars and divide by annual hours, obtaining 123531 records. The records are randomly permuted before local randomization and assigned by arrival index to cohorts of sizes 43235, 37059, and 43237, corresponding to the split $(0.35,0.30,0.35)$. We use log hourly wages, 19 probabilities from 0.05 to 0.95, $\eps=1$, and public bracket $[\log 2,\log 500]$. The gain constant is $0.1\log(500/2)\approx0.552$. The 294 records below this bracket and two above it remain in the outer interval categories.

For calibration, we treat unweighted log wages as independent draws from a continuous working distribution $F$. The target is $F^{-1}(\tau_j)$; the nominal 95\% level refers to repeated independent streams from $F$, with fresh local randomization and multipliers, under \Cref{thm:bootstrap}'s conditions. File quantiles serve as descriptive benchmarks. ACS population inference additionally requires survey weights and the PUMS replicate-weight design.

The released wages have ties: the 123531 log wages take 7260 distinct values, with a largest empirical point mass of 3.2\%. Masses at the 19 empirical target quantiles range from less than 0.01\% to 2.7\%. On the log scale, windows of radius $h_n\approx0.638$ around those quantiles contain 1751--3385 distinct values. The 19 estimated densities range from $0.123$ to $0.507$, with no density clipping or variance fallback. The finite difference therefore averages over many recorded values, while smooth local density remains a working approximation.

We apply the monotone interval refinement in
\Cref{sec:streaming} on the log-wage scale and exponentiate its endpoints
to obtain wage-quantile intervals.  Write $[\wt L_j^I,\wt U_j^I]$ for the refined log-quantile interval.  For $j<\ell$, the ratio of wage
quantiles at $\tau_\ell$ and $\tau_j$ inherits the interval
$[\exp(\wt L_\ell^I-\wt U_j^I),\exp(\wt U_\ell^I-\wt L_j^I)]$.
All such ratio intervals cover whenever the log-quantile band covers.

\begin{table}[!htbp]
\centering
\small
\caption{Hourly-wage quantiles in 2024 dollars. Intervals use nominal 95\% simultaneous calibration under the continuous working model.}
\label{tab:acs-application}
\begin{tabular}{rccc}
\toprule
$\tau$ & Empirical benchmark & Private estimate & 95\% simultaneous interval \\
\midrule
0.10 & 14.40 & 13.44 & [10.85, 16.66] \\
0.25 & 20.50 & 18.96 & [16.27, 22.10] \\
0.50 & 34.17 & 30.71 & [26.72, 35.30] \\
0.75 & 56.52 & 51.50 & [43.69, 60.71] \\
0.90 & 90.30 & 100.37 & [76.54, 131.61] \\
\bottomrule
\end{tabular}
\end{table}

\begin{figure}[!htbp]
\centering
\includegraphics[width=\linewidth]{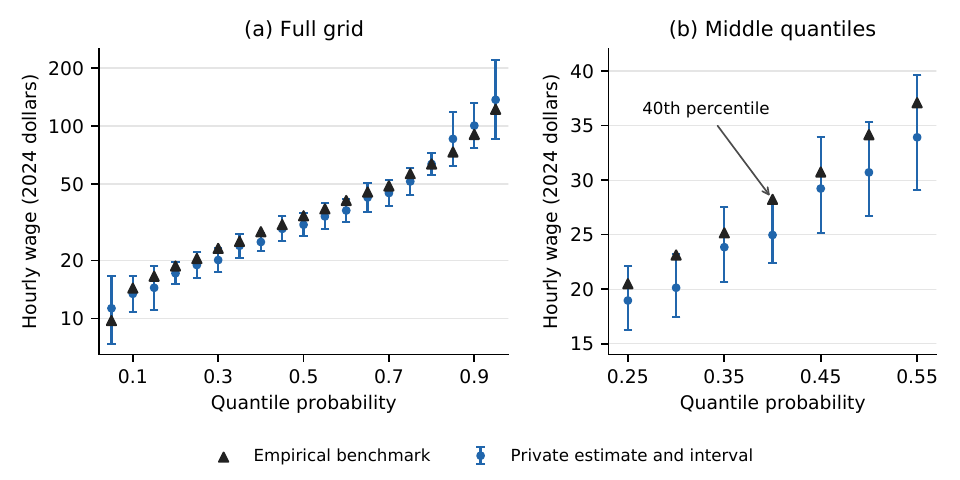}
\caption{Private hourly-wage quantiles and nominal 95\% simultaneous intervals for the 19-point ACS grid in \Cref{sec:application}. The left panel uses a logarithmic wage axis; the right enlarges the middle quantiles on a linear axis. Black triangles are empirical file benchmarks. At the 40th percentile, the benchmark lies above its interval. Calibration targets the continuous working-model quantiles; the plotted benchmarks describe the fixed released file.}
\label{fig:acs-grid}
\end{figure}

The private 90/50 and 50/10 ratios are 3.27 and 2.28, suggesting greater dispersion above the median.  Transforming the simultaneous log-quantile band gives intervals $[2.17,4.93]$ and $[1.60,3.25]$, respectively; the 90/10 ratio is 7.47 with interval $[4.60,12.13]$.  This comparison uses the shared median on the log-wage scale. Writing $Q_\tau=F^{-1}(\tau)$ and $[L_\tau,U_\tau]$ for its simultaneous band endpoints, the contrast $D=Q_{0.90}+Q_{0.10}-2Q_{0.50}$ inherits the interval
\[
 [L_{0.90}+L_{0.10}-2U_{0.50},\;
   U_{0.90}+U_{0.10}-2L_{0.50}].
\]
It contains zero exactly when the rectangular-band intervals for the two log-ratios overlap; exponentiation preserves this overlap. Thus the reported band does not separate upper- and lower-tail dispersion. \Cref{fig:acs-grid} shows all 19 intervals: 18 file quantiles, including all five tabulated benchmarks, lie in the realized band. The exception is the 40th-percentile benchmark $28.23$, whose interval is $[22.41,27.82]$.

\FloatBarrier

\section{Concluding remarks}\label{sec:conclusion}

This paper develops simultaneous quantile inference from one locally private report per user.  The exact covariance of complementary scores yields a variance-minimizing fusion rule without additional privacy cost.  Separating adaptive localization, density estimation, and fresh-score calibration gives a uniform one-step representation: first-order localization error is removed, and the remaining errors can be controlled on the scale of the maximum over the grid.  This supports multiplier bands for polynomially growing grids under the stated conditions.  Fusion improves matched three-stage estimates in the simulations, while comparison with CDF inversion clarifies the sample cost of learning query locations.  The wage application shows how the joint band also supports percentile-ratio comparisons.

Two extensions merit further study.  First, compact channels such as Hadamard response and local hashing \citep{Acharya2019,Wang2017} suggest ways to reduce communication.  Compression changes the dependence among decoded node scores, so the complementary-score covariance and optimal fusion weights would need to be derived anew.  Uniform variance lower bounds and standardized moment bounds would then determine whether the localization rates and Gaussian approximation still permit comparably large grids.  Second, continuously updating query locations and density estimates could make fuller use of an incoming stream.  Time-uniform private quantile inference \citep{Cai2025} offers a starting point, but a growing grid requires a one-step expansion uniform in both time and quantile.  History-dependent scores would replace the present conditionally independent calibration sample, requiring control of accumulated conditional covariances and a Gaussian strong approximation or martingale confidence boundaries.  Combining these bounds with uniform localization control would yield bands valid throughout online updating.

\section*{Acknowledgment}
\par  Qirui Hu's research was supported by National Natural Science Foundation of China (NSFC) (Grant Nos.12601520), the Shanghai Engineering Research Center of Finance Intelligence (Grant No.~19DZ2254600) and by TRR 391 \textit{Spatio-temporal Statistics for the Transition of Energy and Transport} (Project number 520388526) funded by the Deutsche Forschungsgemeinschaft (DFG, German Research Foundation).

\clearpage
\appendix
\section*{Proof organization}\label{app:attribution}

The privacy calculation and fused-score moments are specific to the proposed mechanism.  The Abel decomposition and one-step algebra are established directly.  \Cref{lem:localization} specializes a classical stopped Lyapunov argument for projected stochastic approximation \citep{RobbinsMonro1951,Ruppert1988,PolyakJuditsky1992,KushnerYin2003}.  The Gaussian approximation, multiplier bootstrap, and anti-concentration inputs follow \citet{CCK2013,CCK2017}, with their conditional moment and dimension requirements verified in \Cref{lem:cck}.  The Gaussian mean--sample covariance independence in \Cref{prop:fixedJ} is standard multivariate analysis \citep{Anderson2003}.

\section{Tree-OUE Moments and Fusion}\label{app:moments}

This appendix proves \Cref{prop:fused-moments} from the main text.  Throughout, the public threshold list, its tree, and the anchor are fixed.

\subsection{Node moments}\label{app:node-moments}

Let $I_v=\1\{C(X)\text{ is a descendant leaf of }v\}$.  Condition on $L^\star=\ell(v)$ for the following bit moments; the input bit is $I_v$.  Put
\[
 \eta_v=\frac{Z_v-q_\eps}{p_\eps-q_\eps}.
\]
If $I_v=0$, then $Z_v\sim\operatorname{Bernoulli}(q_\eps)$ and
\begin{align*}
 \E(\eta_v\mid I_v=0)&=0,\\
 \E(\eta_v^2\mid I_v=0)
 &=\frac{q_\eps(1-q_\eps)}{(p_\eps-q_\eps)^2}
 =\frac{4e^\eps}{(e^\eps-1)^2}
 =\nu_\eps.
\end{align*}
If $I_v=1$, then $Z_v\sim\operatorname{Bernoulli}(p_\eps)$ and
\begin{align*}
 \E(\eta_v\mid I_v=1)&=1,\\
 \E(\eta_v^2\mid I_v=1)
 &=1+\frac{p_\eps(1-p_\eps)}{(p_\eps-q_\eps)^2}
 =\nu_\eps+2.
\end{align*}
The last identity follows by direct substitution of $p_\eps=1/2$ and $q_\eps=(e^\eps+1)^{-1}$.

Averaging the sampled-level moments over the uniform $L^\star$ gives
\begin{align}
 \E(\psi_v\mid X)
 &=H_d\Pp\{L^\star=\ell(v)\}\E(\eta_v\mid X,L^\star=\ell(v))
 =I_v,\label{eq:node-first}\\
 \E(\psi_v^2\mid X)
 &=H_d^2\Pp\{L^\star=\ell(v)\}
   \E(\eta_v^2\mid X,L^\star=\ell(v))
 =H_d(\nu_\eps+2I_v).\label{eq:node-second}
\end{align}

If $v\ne w$ are disjoint tree nodes, then
\begin{equation}\label{eq:cross-zero}
 \E(\psi_v\psi_w\mid X)=0.
\end{equation}
Indeed, if their levels differ, the two level indicators cannot both equal one.  If their levels agree, their unary bits are conditionally independent and at most one input bit equals one.  The centered decoder for every zero input has conditional mean zero, so the product has conditional mean zero.

Finally, since $|Z_v-q_\eps|\le1-q_\eps$ and
\[
 \frac{1-q_\eps}{p_\eps-q_\eps}
 =\frac{2e^\eps}{e^\eps-1},
\]
we have
\[
 |\psi_v|\le\frac{2H_de^\eps}{e^\eps-1}.
\]

\subsection{Prefix, suffix, and fusion}\label{app:fusion-proof}

Let $p_r=F(x_{(r)})$.  Exactly one node in $\cD_r^-$ contains the realized leaf when $X\le x_{(r)}$, and none does otherwise.  Summing \eqref{eq:node-first} gives
\[
 \E(A_r\mid X)=\1\{X\le x_{(r)}\},
 \qquad
 \E A_r=p_r.
\]
Using \eqref{eq:node-second} and \eqref{eq:cross-zero},
\begin{align*}
 \E A_r^2
 &=\sum_{v\in\cD_r^-}\E\psi_v^2
 =H_dw_r^-\nu_\eps+2H_d
   \sum_{v\in\cD_r^-}\Pp(I_v=1)\\
 &=H_dw_r^-\nu_\eps+2H_dp_r.
\end{align*}
Therefore
\[
 \Var(A_r)=H_dw_r^-\nu_\eps+2H_dp_r-p_r^2.
\]

Write $S_r=\sum_{v\in\cD_r^+}\psi_v$, so $B_r=1-S_r$.  The same argument gives
\[
 \E S_r=1-p_r,
 \qquad
 \E S_r^2=H_dw_r^+\nu_\eps+2H_d(1-p_r),
\]
and hence
\[
 \Var(B_r)=\Var(S_r)
 =H_dw_r^+\nu_\eps+2H_d(1-p_r)-(1-p_r)^2.
\]
Every prefix node is disjoint from every suffix node, so \eqref{eq:cross-zero} yields $\E(A_rS_r)=0$.  It follows that
\begin{align*}
 \Cov(A_r,B_r)
 &=\Cov(A_r,1-S_r)
 =-\{\E(A_rS_r)-\E A_r\E S_r\}
 =p_r(1-p_r).
\end{align*}

For an arbitrary scalar $\lambda$, the estimator
$\lambda A_r+(1-\lambda)B_r$ is unbiased, with variance
\begin{align}
 \Var\{\lambda A_r+(1-\lambda)B_r\}
 ={}&\lambda^2\Var(A_r)+(1-\lambda)^2\Var(B_r)
 +2\lambda(1-\lambda)p_r(1-p_r),\label{eq:quadvar}
\end{align}
Set $C_r=p_r(1-p_r)$ and
\[
 a=a_r^-(p_r)=\Var(A_r)-C_r,
 \qquad
 b=a_r^+(p_r)=\Var(B_r)-C_r.
\]
Completing the square gives
\begin{equation}\label{eq:complete-square}
 \Var\{\lambda A_r+(1-\lambda)B_r\}
 =C_r+a\lambda^2+b(1-\lambda)^2
 =C_r+\frac{ab}{a+b}+(a+b)\left(\lambda-\frac{b}{a+b}\right)^2.
\end{equation}
Both $a$ and $b$ are strictly positive.  The minimizer is $b/(a+b)=\lambda_r^\circ(p_r)$ and the minimum is \eqref{eq:fused-min-var}.  Since minimizing over $[0,1]$ cannot be worse than evaluating at zero or one, the variance dominance follows.

The sum
$D_r=a_r^-(u)+a_r^+(u)=H_d(w_r^-+w_r^+)\nu_\eps+2H_d-1$
does not depend on $u$, and
\[
 \lambda_r^\circ(u)-\lambda_r^\circ(p_r)
 =-\frac{(2H_d-1)(u-p_r)}{D_r}.
\]
Substitution into \eqref{eq:complete-square} gives
\eqref{eq:fusion-excess}.  Since $D_r\ge2H_d-1$, it also gives
\eqref{eq:fusion-excess-bound}.  Finally, at the sampled level each
canonical decomposition contributes at most one node score.  Hence
$|A_r|\le2H_de^\eps/(e^\eps-1)$ and
$|B_r|\le1+2H_de^\eps/(e^\eps-1)$.  Convexity of the fused score proves
the claimed envelope.  \qed

\section{Privacy Proof}\label{app:privacy}

\begin{proof}[Sequential privacy guarantee]
Fix a public threshold list and a sampled level $\ell$.  Two possible interval categories produce unary vectors that are either identical or differ in exactly two coordinates: one coordinate changes from one to zero and another from zero to one.  For one changed pair and any released bit pair $(z_1,z_2)$, the largest likelihood ratio is
\[
 \frac{p_\eps(1-q_\eps)}{q_\eps(1-p_\eps)}
 =\frac{1-q_\eps}{q_\eps}
 =e^\eps.
\]
All unchanged bit factors cancel.  Thus the conditional OUE release at level $\ell$ is $\eps$-LDP.  The level is sampled independently of the private category, so mixing over levels preserves the same bound.

Now consider the sequential transcript $Z_{1:n}$.  Let two data streams differ only at user $i$.  For a fixed realized transcript, the chain rule factorizes its likelihood into conditional kernels.  The factors before $i$ coincide.  At $i$, the public query is the same function of the common preceding transcript and the conditional likelihood ratio is at most $e^{\eps_n}$.  Conditional on the realized report $Z_i$ and the subsequent common transcript prefix, all later public queries are again the same deterministic functions of that prefix.  Because later users' raw values agree, every later conditional kernel cancels in the likelihood ratio.  Hence the complete transcript ratio is at most $e^{\eps_n}$.

All server computations after receipt of the reports---sorting, score decoding, projection, averaging, density estimation, multiplier generation, and interval refinement---are measurable post-processing and cannot increase the privacy loss.  Finally, a user belongs to exactly one cohort, so there is no across-stage composition for that user.  This proves the privacy guarantee.
\end{proof}

\section{Stage-1 Localization and Full-Average Expansion}\label{app:stage1}

We first record the filtration and a localization lemma.  For Stage 1, let $\cF_t$ contain all observations only through their released reports up to time $t$, together with all public randomization.  Put
\[
 e_{t,j}=q_{t,j}-Q_{j,n},
 \qquad
 \zeta_{t,j}=G_{t,j}^{(1)}-F(q_{t-1,j}).
\]
Then
\begin{equation}\label{eq:mds-properties}
 \E(\zeta_{t,j}\mid\cF_{t-1})=0,
 \qquad
 |\zeta_{t,j}|\le 2B_{1n}.
\end{equation}
The bound uses $0\le F\le1$ and $|G_{t,j}^{(1)}|\le B_{1n}$, enlarging an inessential universal constant.

\begin{lemma}[Uniform localization of the projected recursion]\label{lem:localization}
Under \Cref{ass:iid,ass:identification,ass:gain}, there is a constant $C$ depending only on $(\mu,L_F,\gamma_0,a)$ such that, for a localization index satisfying \eqref{eq:bn1}--\eqref{eq:bn3}, an event $\cE_{\mathrm{loc},n}$ has probability tending to one and, on this event,
\begin{enumerate}[label=(\roman*),leftmargin=2em]
\item $|e_{t,j}|\le\delta_0$ for every $j\le k_n$ and $b_n\le t\le n_1$;
\item the scalar projection in \eqref{eq:sgd-update} is inactive for $b_n<t\le n_1$;
\item uniformly over $j\le k_n$ and $b_n\le t\le n_1$,
\begin{equation}\label{eq:iterate-bound}
 |e_{t,j}|
 \le C\left[
 e^{-\frac\mu2\sum_{s=b_n+1}^t\gamma_s}|e_{b_n,j}|
 +B_{1n}\{\sqrt{\gamma_t\ell_n}+\gamma_t\ell_n\}
 \right].
\end{equation}
\end{enumerate}
A sufficient primitive condition for selecting such a $b_n$ is
\[
 B_{1n}^2\ell_n\omega_n=o(n_1^a),
 \qquad
 b_n\asymp
 \left(B_{1n}^2\ell_n\omega_n\right)^{1/a}
 +\left(\log n\right)^{1/(1-a)},
 \qquad b_n=o(n_1),
\]
where $\omega_n\to\infty$ may diverge arbitrarily slowly and the multiplicative constants are chosen large enough.
\end{lemma}

\begin{proof}
We use deterministic contraction weights so that every innovation is
multiplied by a predictable coefficient.

\textit{Step 1: a stopped Lyapunov inequality.}
Projection onto $[L,U]$ is nonexpansive and fixes $Q_{j,n}$.  With
$h_{t-1,j}=F(q_{t-1,j})-\tau_{j,n}$,
\[
 e_{t,j}^2
 \le \left[e_{t-1,j}-\gamma_t\{h_{t-1,j}+\zeta_{t,j}\}\right]^2.
\]
\begingroup
The signed secant condition gives
$\mu\le h_{t-1,j}/e_{t-1,j}\le L_F$ whenever $e_{t-1,j}\ne0$.
Since $\gamma_tL_F\le1/2$,
\[
 |e_{t-1,j}-\gamma_th_{t-1,j}|
 \le(1-\mu\gamma_t)|e_{t-1,j}|
 \le |e_{t-1,j}|.
\]
These inequalities also hold when $e_{t-1,j}=0$, since then $h_{t-1,j}=0$.
Squaring the first inequality and using
$(1-\mu\gamma_t)^2\le1-\mu\gamma_t$ gives the drift contraction.
\endgroup
After expanding the square, define
\begin{equation}\label{eq:lyap-increment}
 \Delta M_{t,j}
 =-2\gamma_t(e_{t-1,j}-\gamma_th_{t-1,j})\zeta_{t,j}
  +\gamma_t^2\{\zeta_{t,j}^2-
  \E(\zeta_{t,j}^2\mid\cF_{t-1})\}.
\end{equation}
Then $\E(\Delta M_{t,j}\mid\cF_{t-1})=0$.  The secant bounds, the gain restriction, and \eqref{eq:mds-properties} give
\begin{align}
 e_{t,j}^2
 &\le(1-\mu\gamma_t)e_{t-1,j}^2
   +C\gamma_t^2B_{1n}^2+\Delta M_{t,j},
 \label{eq:lyap-recursion}\\
 |\Delta M_{t,j}|
 &\le C\gamma_tB_{1n}
       (|e_{t-1,j}|+\gamma_tB_{1n}),
 \label{eq:lyap-increment-bound}\\
 \E(\Delta M_{t,j}^2\mid\cF_{t-1})
 &\le C\gamma_t^2B_{1n}^2
       (e_{t-1,j}^2+\gamma_t^2B_{1n}^2).
 \label{eq:lyap-qv-bound}
\end{align}
In particular, taking expectations and iterating \eqref{eq:lyap-recursion} yields
\[
 \E e_{t,j}^2
 \le (U-L)^2\exp\left(-\mu\sum_{s=1}^t\gamma_s\right)
 +CB_{1n}^2\gamma_t.
\]

For the uniform probability bound, fix a deterministic $r<t$.  Iterating
\eqref{eq:lyap-recursion} from $r$ to $t$ involves only the deterministic
weights
\[
 \rho_{s,t}=\prod_{u=s+1}^t(1-\mu\gamma_u),
 \qquad \rho_{t,t}=1.
\]
Because $\mu\gamma_u\le1/2$,
\[
 \rho_{s,t}
 \le \exp\left(-\mu\sum_{u=s+1}^t\gamma_u\right).
\]
For every fixed $q\ge1$ and $m>1$,
\begin{equation}\label{eq:weight-convolution-general}
 \sum_{s=r+1}^t\rho_{s,t}^q\gamma_s^m
 \le C_{q,m}\gamma_t^{m-1},
 \qquad
 \max_{r<s\le t}\rho_{s,t}^q\gamma_s^m
 \le C_{q,m}\gamma_t^m.
\end{equation}
The maximum inequality in \eqref{eq:weight-convolution-general} also
holds for every $m>0$.

To verify these bounds, split the index set at $t/2$.  If $s>t/2$,
then $\gamma_s\le2^a\gamma_t$ and
\[
 \rho_{s,t}^q
 \le \exp\{-q\mu(t-s)\gamma_t\}.
\]
Consequently,
\[
 \sum_{t/2<s\le t}\rho_{s,t}^q\gamma_s^m
 \le C\gamma_t^m\sum_{h=0}^{\infty}e^{-q\mu h\gamma_t}
 \le C\gamma_t^{m-1}.
\]
If $s\le t/2$, then
\[
 \rho_{s,t}^q
 \le \exp\left\{-q\mu
 \sum_{u=\lfloor t/2\rfloor+1}^t\gamma_u\right\}
 \le \exp(-ct^{1-a}),
\]
and hence
\[
 \sum_{r<s\le t/2}\rho_{s,t}^q\gamma_s^m
 \le Ct e^{-ct^{1-a}}
 \le C\gamma_t^{m-1}.
\]
The same split proves the maximum bound.  In particular,
\[
 \sum_{s=r+1}^t\rho_{s,t}\gamma_s^2\le C\gamma_t,
 \qquad
 \max_{r<s\le t}\rho_{s,t}\gamma_s\le C\gamma_t.
\]
For $x\ge1$, define the first crossing time
\[
 \tau=\inf\left\{s\in\{r+1,\ldots,n_1\}:e_{s,j}^2>
 C_0\left[\rho_{r,s}e_{r,j}^2+B_{1n}^2
 \{\gamma_sx+\gamma_s^2x^2\}\right]\right\},
 \qquad \inf\varnothing=n_1+1.
\]
For a fixed candidate terminal time $t>r$, put
\[
 \widetilde M_{u,t,j}
 =\sum_{s=r+1}^{u}\1\{\tau\ge s\}\rho_{s,t}\Delta M_{s,j},
 \qquad r\le u\le t.
\]
Because $\{\tau\ge s\}\in\cF_{s-1}$ and $\rho_{s,t}$ is deterministic,
$(\widetilde M_{u,t,j})_{u=r}^t$ is a martingale.  Its predictable
quadratic variation and largest increment are
\begin{align*}
 V_{r,t,j}
 &=\sum_{s=r+1}^t
 \E\!\left[
 \{\1(\tau\ge s)\rho_{s,t}\Delta M_{s,j}\}^2
 \mid\cF_{s-1}\right],\\
 L_{r,t,j}
 &=\max_{r<s\le t}
 |\1(\tau\ge s)\rho_{s,t}\Delta M_{s,j}|.
\end{align*}
On $\{\tau\ge s\}$, if $s\ge r+2$ the absence of an earlier crossing
gives the boundary at $s-1$; if $s=r+1$, the same bound follows directly
from $e_{s-1,j}=e_{r,j}$.  Thus, in both cases,
\[
 e_{s-1,j}^2\le C\left[
 \rho_{r,s-1}e_{r,j}^2+
 B_{1n}^2\{\gamma_sx+\gamma_s^2x^2\}\right].
\]
Combining this bound with
\eqref{eq:lyap-increment-bound}--\eqref{eq:lyap-qv-bound} gives
\begin{align}
 V_{r,t,j}
 &\le C\left[
 B_{1n}^2e_{r,j}^2\rho_{r,t}\gamma_t
 +B_{1n}^4\{\gamma_t^2x+\gamma_t^3x^2\}
 \right],\label{eq:stopped-qv}\\
 L_{r,t,j}
 &\le C\left[
 B_{1n}|e_{r,j}|\rho_{r,t}^{1/2}\gamma_t
 +B_{1n}^2\{\gamma_t^{3/2}x^{1/2}+\gamma_t^2x\}
 \right].\label{eq:stopped-increment}
\end{align}
Here are the summations behind these two displays.  The product identity
and the gain restriction give
\[
 \rho_{s,t}\rho_{r,s-1}
 =\frac{\rho_{r,t}}{1-\mu\gamma_s},
 \qquad
 (1-\mu\gamma_s)^{-1}\le2.
\]
Therefore,
\begin{equation}\label{eq:weight-init-sum}
 \sum_{s=r+1}^t
 \rho_{s,t}^2\gamma_s^2\rho_{r,s-1}
 \le C\rho_{r,t}\sum_{s=r+1}^t
 \rho_{s,t}\gamma_s^2
 \le C\rho_{r,t}\gamma_t.
\end{equation}
\[
 \sum_{s=r+1}^t\rho_{s,t}^2\gamma_s^3
 \le C\gamma_t^2,
 \qquad
 \sum_{s=r+1}^t\rho_{s,t}^2\gamma_s^4
 \le C\gamma_t^3.
\]
The corresponding maximum bounds are
\begin{align}
 \max_{r<s\le t}
 \rho_{s,t}\gamma_s\rho_{r,s-1}^{1/2}
 &\le C\rho_{r,t}^{1/2}\gamma_t,
 \label{eq:weight-init-max}\\
 \max_{r<s\le t}\rho_{s,t}\gamma_s^{3/2}
 &\le C\gamma_t^{3/2},
 \qquad
 \max_{r<s\le t}\rho_{s,t}\gamma_s^2
 \le C\gamma_t^2.
 \label{eq:weight-noise-max}
\end{align}
Indeed, the square of the left-hand side of
\eqref{eq:weight-init-max} is at most
\[
 C\rho_{r,t}\rho_{s,t}\gamma_s^2
 \le C\rho_{r,t}\gamma_t^2
\]
by \eqref{eq:weight-convolution-general}.  Equations
\eqref{eq:weight-init-sum}--\eqref{eq:weight-noise-max} prove
\eqref{eq:stopped-qv}--\eqref{eq:stopped-increment}.

Let $\overline V_{r,t,j}$ and $\overline L_{r,t,j}$ denote the respective
right-hand sides of \eqref{eq:stopped-qv} and
\eqref{eq:stopped-increment}.  They are $\cF_r$-measurable upper bounds for
the predictable variation and the increment size.  Freedman's inequality
now yields
\[
 \Pp\!\left\{
 \widetilde M_{t,t,j}>
 \sqrt{2x\overline V_{r,t,j}}+\frac{x\overline L_{r,t,j}}3
 \,\middle|\,\cF_r\right\}\le e^{-x}.
\]
The initialization terms in the Freedman threshold satisfy, for every
fixed $\eta>0$,
\begin{align*}
 B_{1n}|e_{r,j}|\sqrt{\rho_{r,t}\gamma_tx}
 &\le
 \eta\rho_{r,t}e_{r,j}^2+
 C_\eta B_{1n}^2\gamma_tx,\\
 xB_{1n}|e_{r,j}|\rho_{r,t}^{1/2}\gamma_t
 &\le
 \eta\rho_{r,t}e_{r,j}^2+
 C_\eta B_{1n}^2\gamma_t^2x^2.
\end{align*}
Since $(\gamma_tx)^{3/2}\le\gamma_tx+(\gamma_tx)^2$,
\[
 \sqrt{2x\overline V_{r,t,j}}+
 \frac{x\overline L_{r,t,j}}3
 \le C\left[
 \rho_{r,t}e_{r,j}^2+
 B_{1n}^2\{\gamma_tx+\gamma_t^2x^2\}\right].
\]
On $\{\tau=t\}$ all indicators in $\widetilde M_{t,t,j}$ equal one,
and iteration of \eqref{eq:lyap-recursion} gives
\[
 e_{t,j}^2
 \le \rho_{r,t}e_{r,j}^2+
 CB_{1n}^2\gamma_t+\widetilde M_{t,t,j}.
\]
Since $x\ge1$, the middle term is absorbed by
$CB_{1n}^2\gamma_tx$.  Choosing $C_0$ larger than the constants above
therefore implies
\[
 \{\tau=t\}
 \subseteq
 \left\{
 \widetilde M_{t,t,j}>
 \sqrt{2x\overline V_{r,t,j}}+
 \frac{x\overline L_{r,t,j}}3
 \right\}.
\]
Summing its probability
over the possible first crossing times proves
\begin{equation}\label{eq:stopped-lyap-prob}
 \Pp\left(
 \exists t\in\{r+1,\ldots,n_1\}:\
 e_{t,j}^2>C\left[
 \rho_{r,t}e_{r,j}^2+B_{1n}^2
 \{\gamma_tx+\gamma_t^2x^2\}\right]
 \ \middle|\ \cF_r\right)
 \le 2(n_1-r)e^{-x}.
\end{equation}
This argument is the bounded-noise, one-dimensional stopped Lyapunov
specialization of the projected stochastic-approximation stability proof;
\eqref{eq:lyap-increment}--\eqref{eq:stopped-lyap-prob} display the
martingale and its predictable weights explicitly.

\textit{Step 2: uniform entry and inactive projection.}
Apply \eqref{eq:stopped-lyap-prob} with $r=0$ and $x=c\ell_n$, and take a
union bound over $j\le k_n$.  Because
$\rho_{0,t}\le\exp(-\mu\sum_{s=1}^t\gamma_s)$, conditions
\eqref{eq:bn2}--\eqref{eq:bn3} imply, for all sufficiently large $n$,
that $|e_{t,j}|\le\delta_0$ simultaneously for $t\ge b_n$ and $j\le k_n$.
If $t>b_n$, the unprojected displacement is bounded by
\[
 \gamma_t\{|h_{t-1,j}|+|\zeta_{t,j}|\}
 \le C\gamma_{b_n}B_{1n}=o(\delta_0).
\]
Since every $Q_{j,n}$ is at least $3\delta_0$ from the endpoints of the
public bracket, the unprojected update remains in $[L,U]$.  Projection is
therefore inactive after $b_n$.

Finally apply \eqref{eq:stopped-lyap-prob} conditionally with $r=b_n$.
The deterministic weights satisfy
\[
 \rho_{b_n,t}^{1/2}\le
 \exp\left(-\frac{\mu}{2}\sum_{s=b_n+1}^t\gamma_s\right).
\]
Taking square roots and union bounding over $(t,j)$ gives
\eqref{eq:iterate-bound} and completes the three assertions.

\textit{Step 3: a concrete $b_n$.}
For $\gamma_t=\gamma_0t^{-a}$,
\[
 \sum_{s=1}^{b_n}\gamma_s\asymp b_n^{1-a},
 \qquad
 \gamma_{b_n}\asymp b_n^{-a}.
\]
The displayed order for $b_n$ simultaneously makes the initialization factor polynomially small and $B_{1n}\sqrt{\gamma_{b_n}\ell_n}=O(\omega_n^{-1/2})=o(1)$.  The corresponding requirement $B_{1n}^2\ell_n\omega_n=o(n_1^a)$ permits $b_n=o(n_1)$.  This completes the proof.
\end{proof}

\begin{lemma}[Abel decomposition for the full average]\label{lem:abel}
Suppose \Cref{ass:smooth} holds.  On $\cE_{\mathrm{loc},n}$, for each $j$,
\begin{equation}\label{eq:abel-identity}
\begin{split}
 f(Q_{j,n})\frac1{n_1}\sum_{t=b_n+1}^{n_1}e_{t-1,j}
 ={}&-\frac1{n_1}\sum_{t=b_n+1}^{n_1}\zeta_{t,j}
 -\frac1{n_1}\sum_{t=b_n+1}^{n_1}\mathcal R_{t-1,j}\\
 &+\frac1{n_1}\left\{
 \frac{e_{b_n,j}}{\gamma_{b_n+1}}
 -\frac{e_{n_1,j}}{\gamma_{n_1}}
 +\sum_{t=b_n+1}^{n_1-1}e_{t,j}
 \left(\frac1{\gamma_{t+1}}-\frac1{\gamma_t}\right)
 \right\},
\end{split}
\end{equation}
where
\[
 \mathcal R_{t-1,j}
 =F(Q_{j,n}+e_{t-1,j})-F(Q_{j,n})-f(Q_{j,n})e_{t-1,j}
\]

satisfies
\begin{equation}\label{eq:taylor-holder}
 |\mathcal R_{t-1,j}|
 \le\frac{L_f}{1+\beta}|e_{t-1,j}|^{1+\beta}.
\end{equation}
\end{lemma}

\begin{proof}
Projection is inactive after $b_n$, so
\[
 e_{t,j}=e_{t-1,j}-\gamma_t
 \{F(Q_{j,n}+e_{t-1,j})-F(Q_{j,n})+\zeta_{t,j}\}.
\]
Rearrange this equality to obtain
\[
 f(Q_{j,n})e_{t-1,j}
 =\frac{e_{t-1,j}-e_{t,j}}{\gamma_t}
 -\zeta_{t,j}-\mathcal R_{t-1,j}.
\]
Summing from $b_n+1$ to $n_1$ gives the first two terms in \eqref{eq:abel-identity}.  For the difference term,
\begin{align*}
 \sum_{t=b_n+1}^{n_1}\frac{e_{t-1,j}-e_{t,j}}{\gamma_t}
 ={}&\frac{e_{b_n,j}}{\gamma_{b_n+1}}
 -\frac{e_{n_1,j}}{\gamma_{n_1}}
 +\sum_{t=b_n+1}^{n_1-1}e_{t,j}
 \left(\frac1{\gamma_{t+1}}-\frac1{\gamma_t}\right),
\end{align*}
which is summation by parts.  Finally,
\begin{align*}
 \mathcal R_{t-1,j}
 &=e_{t-1,j}\int_0^1
 \{f(Q_{j,n}+se_{t-1,j})-f(Q_{j,n})\}\,ds,
\end{align*}
so \Cref{ass:smooth} yields \eqref{eq:taylor-holder}.
\end{proof}

\begin{lemma}[Uniform bounds for the terms in \eqref{eq:abel-identity}]\label{lem:abel-bounds}
Under \Cref{ass:iid,ass:smooth,ass:identification,ass:gain}, on events
whose probabilities tend to one,
\begin{align}
 \max_{j\le k_n}\left|\frac1{n_1}\sum_{t=b_n+1}^{n_1}\zeta_{t,j}\right|
 &\le C B_{1n}\sqrt{\frac{\ell_n}{n_1}},\label{eq:mart-bound}\\
 \max_{j\le k_n}\frac1{n_1}
 \left|\frac{e_{b_n,j}}{\gamma_{b_n+1}}-\frac{e_{n_1,j}}{\gamma_{n_1}}
 +\sum_{t=b_n+1}^{n_1-1}e_{t,j}
 \left(\frac1{\gamma_{t+1}}-\frac1{\gamma_t}\right)\right|
 &\le C B_{1n}n_1^{a/2-1}\sqrt{\ell_n}
 +o(r_{1n}),\label{eq:abel-bound}\\
 \max_{j\le k_n}\frac1{n_1}
 \sum_{t=b_n+1}^{n_1}|\mathcal R_{t-1,j}|
 &\le C B_{1n}^{1+\beta}n_1^{-a(1+\beta)/2}
 \ell_n^{(1+\beta)/2}+o(r_{1n}).
 \label{eq:nonlinear-bound}
\end{align}
The displayed $o(r_{1n})$ terms collect the exponentially decaying
initialization contribution.
\end{lemma}

\begin{proof}
For \eqref{eq:mart-bound}, apply Bernstein's martingale inequality to the bounded differences $\zeta_{t,j}$ and use a union bound over $j$.  The resulting bound is
\[
 C B_{1n}\left\{\sqrt{\frac{\ell_n}{n_1}}+\frac{\ell_n}{n_1}\right\};
\]
By \eqref{eq:bn3}, $b_n<n_1$, and
$\gamma_{b_n}=\gamma_0b_n^{-a}$,
$B_{1n}^2\ell_n=o(b_n^a)=o(n_1^a)$.  Since $a<1$, this implies
$\ell_n/n_1=o(1)$, so the second term is dominated by the first.

For \eqref{eq:abel-bound}, write $b=b_n$ and put
\[
 D_{b,t}
 =\exp\left(-\frac{\mu}{2}\sum_{s=b+1}^t\gamma_s\right),
 \qquad D_{b,b}=1.
\]
The same near--far split used in
\eqref{eq:weight-convolution-general} gives, for every fixed $q>0$,
\begin{align}
 \sup_{t\ge b}D_{b,t}
 \left(\frac{\gamma_b}{\gamma_t}\right)^q
 &\le C_q,
 \label{eq:abel-propagation-sup}\\
 \sum_{t=b}^{N}D_{b,t}^q
 &\le C_q\gamma_b^{-1},
 \label{eq:abel-propagation-sum}\\
 \frac{D_{b,N}}{\gamma_N}
 +\sum_{t=b+1}^{N-1}D_{b,t}t^{a-1}
 &\le C\gamma_b^{-1}.
 \label{eq:abel-propagation-weighted}
\end{align}
Indeed, when $b<t\le2b$,
\[
 \sum_{s=b+1}^t\gamma_s\ge c(t-b)\gamma_b,
\]
so the corresponding sums are dominated by a geometric series with
effective length $O(\gamma_b^{-1})$.  When $t>2b$,
\[
 \sum_{s=b+1}^t\gamma_s
 \ge c(t^{1-a}-b^{1-a}),
\]
and the remaining tail is exponentially small.

Applying \eqref{eq:stopped-lyap-prob} with $r=0$ and $x=c\ell_n$ at
time $b$ gives
\[
 |e_{b,j}|\le d_{0n}
 +CB_{1n}\{\sqrt{\gamma_b\ell_n}+\gamma_b\ell_n\},
 \qquad
 d_{0n}=(U-L)e^{-\frac\mu2\sum_{s=1}^{b}\gamma_s}
 =o(r_{1n}).
\]
Combining this display with \eqref{eq:iterate-bound} and
\eqref{eq:abel-propagation-sup} yields
\begin{equation}\label{eq:iterate-bound-separated}
 |e_{t,j}|
 \le CD_{b,t}d_{0n}
 +CB_{1n}\{\sqrt{\gamma_t\ell_n}+\gamma_t\ell_n\}.
\end{equation}
Here we used, respectively, $q=1/2$ and $q=1$ in
\eqref{eq:abel-propagation-sup} to propagate the two stochastic terms at
time $b$.  By \eqref{eq:bn3}, $B_{1n}\ge1$, and monotonicity of the gain,
$\gamma_t\ell_n\le\gamma_b\ell_n=o(1)$ for $t\ge b$; hence the second
term in \eqref{eq:iterate-bound-separated} is bounded by
$CB_{1n}t^{-a/2}\sqrt{\ell_n}$.

The contribution of the deterministic initialization to the Abel bracket
is, by \eqref{eq:abel-propagation-weighted},
\[
 \frac{Cd_{0n}}{n_1}\left[
 \gamma_b^{-1}
 +\frac{D_{b,n_1}}{\gamma_{n_1}}
 +\sum_{t=b+1}^{n_1-1}D_{b,t}t^{a-1}
 \right]
 \le \frac{Cd_{0n}}{n_1\gamma_b}=o(r_{1n}),
\]
because $b=o(n_1)$ and $a<1$ imply
$n_1\gamma_b\asymp n_1b^{-a}\to\infty$.
For the remaining terms,
\[
 \frac1{\gamma_{t+1}}-\frac1{\gamma_t}
 =\gamma_0^{-1}\{(t+1)^a-t^a\}
 \le Ct^{a-1},
\]
and therefore
\[
 \sum_{t=b+1}^{n_1-1}|e_{t,j}|
 \left(\frac1{\gamma_{t+1}}-\frac1{\gamma_t}\right)
 \le CB_{1n}\sqrt{\ell_n}
 \sum_{t\le n_1}t^{a/2-1}
 +o(n_1r_{1n})
 \le CB_{1n}\sqrt{\ell_n}\,n_1^{a/2}
 +o(n_1r_{1n}).
\]
The two endpoint terms obey the same bound:
\[
 \frac{|e_{b,j}|}{\gamma_{b+1}}
 +\frac{|e_{n_1,j}|}{\gamma_{n_1}}
 \le CB_{1n}\sqrt{\ell_n}\,n_1^{a/2}
 +o(n_1r_{1n}).
\]
Dividing by $n_1$ proves \eqref{eq:abel-bound}.

For \eqref{eq:nonlinear-bound}, combine
\eqref{eq:taylor-holder} and \eqref{eq:iterate-bound-separated}.  Since
$(x+y)^{1+\beta}\le2^\beta(x^{1+\beta}+y^{1+\beta})$,
\[
 |e_{t,j}|^{1+\beta}
 \le CD_{b,t}^{1+\beta}d_{0n}^{1+\beta}
 +CB_{1n}^{1+\beta}\gamma_t^{(1+\beta)/2}
 \ell_n^{(1+\beta)/2}.
\]
The stochastic sum is bounded by
\begin{align*}
 \frac{B_{1n}^{1+\beta}\ell_n^{(1+\beta)/2}}{n_1}
 \sum_{t\le n_1}t^{-a(1+\beta)/2}
 \le C B_{1n}^{1+\beta}\ell_n^{(1+\beta)/2}
 n_1^{-a(1+\beta)/2},
\end{align*}
where $a(1+\beta)/2<1$.  By
\eqref{eq:abel-propagation-sum}, the initialization contribution is at
most
\[
 \frac{Cd_{0n}^{1+\beta}}{n_1\gamma_b}=o(r_{1n}),
\]
because $d_{0n}=o\{\min(r_{1n},1)\}$ and
$n_1\gamma_b\to\infty$.  This proves all three bounds.
\end{proof}

\begin{proof}[Proof of \Cref{thm:preliminary}]
Write $f_j=f(Q_{j,n})$.  Since
\[
 \sum_{t=b_n+1}^{n_1}e_{t,j}
 =\sum_{t=b_n+1}^{n_1}e_{t-1,j}
 +e_{n_1,j}-e_{b_n,j},
\]
\Cref{lem:abel} gives the exact expansion
\[
 f_j(q_j^0-Q_{j,n})
 =-\frac1{n_1}\sum_{t=b_n+1}^{n_1}\zeta_{t,j}
 +R_{1n,j},
\]
where
\begin{align}
 R_{1n,j}
 ={}&\frac{f_j}{n_1}\sum_{t=1}^{b_n}e_{t,j}
 +\frac{f_j}{n_1}(e_{n_1,j}-e_{b_n,j})\notag\\
 &-\frac1{n_1}\sum_{t=b_n+1}^{n_1}\mathcal R_{t-1,j}
 +\frac1{n_1}\left\{
 \frac{e_{b_n,j}}{\gamma_{b_n+1}}
 -\frac{e_{n_1,j}}{\gamma_{n_1}}
 +\sum_{t=b_n+1}^{n_1-1}e_{t,j}
 \left(\frac1{\gamma_{t+1}}-\frac1{\gamma_t}\right)
 \right\}.
 \label{eq:R1n-exact}
\end{align}
The first term is at most $f_+(U-L)b_n/n_1$.  By
\eqref{eq:iterate-bound-separated},
\[
 \frac{|e_{n_1,j}|+|e_{b_n,j}|}{n_1}
 \le \frac{Cd_{0n}}{n_1}
 +CB_{1n}n_1^{a/2-1}\sqrt{\ell_n}
 =o(r_{1n})
 +O\!\left(B_{1n}n_1^{a/2-1}\sqrt{\ell_n}\right).
\]
The last two terms in \eqref{eq:R1n-exact} are bounded by
\eqref{eq:nonlinear-bound} and \eqref{eq:abel-bound}.  Together with
\eqref{eq:mart-bound}, these are exactly the four orders in
\eqref{eq:r1n}.  Since $f_j\ge f_-$ uniformly, this proves
\eqref{eq:prelim-rate} and \eqref{eq:stage1-expansion}.
\end{proof}

\begin{proof}[Proof of \Cref{prop:full-tail}]
For $b>0$,
\[
 \ol q_{n_1,j}
 =\frac{b}{n_1}\left(\frac1b\sum_{t=1}^bq_{t,j}\right)
 +\frac{n_1-b}{n_1}\ol q_{n_1,j}^{(b)}.
\]
Therefore
\[
 \ol q_{n_1,j}^{(b)}-\ol q_{n_1,j}
 =\frac{b}{n_1}
 \left\{\ol q_{n_1,j}^{(b)}-
 \frac1b\sum_{t=1}^bq_{t,j}\right\}.
\]
Both terms in braces lie in $[L,U]$, so the absolute difference is at most $(U-L)b/n_1$.  The case $b=0$ is immediate.
\end{proof}

\section{Private Density Estimation and the One-Step Expansion}\label{app:onestep}

\begin{lemma}[Conditional concentration of Stage-2 CDF scores]\label{lem:stage2-concentration}
Under \Cref{ass:iid}, conditional on Stage 1, with conditional
probability tending to one,
\[
 \max_{j\le k_n}\left\{
 \left|\ol G_{j,-}^{(2)}-F(q_j^0-h_n)\right|,
 \left|\ol G_{j,+}^{(2)}-F(q_j^0+h_n)\right|
 \right\}
 \le C B_{2n}
 \left\{\sqrt{\frac{\ell_n}{n_2}}+\frac{\ell_n}{n_2}\right\}.
\]
\end{lemma}

\begin{proof}
Once Stage 1 is fixed, the $2k_n$ thresholds, their ranks, and all anchors are deterministic.  The Stage-2 user vectors are conditionally i.i.d.  By \Cref{prop:fused-moments}, each coordinate is unbiased and bounded in absolute value by $B_{2n}$.  Therefore the centered coordinate is bounded by $2B_{2n}$.  For either sign, Bernstein's inequality gives, for any $x>0$,
\[
 \Pp\left(
 |\ol G_{j,\pm}^{(2)}-F(q_j^0\pm h_n)|
 >CB_{2n}\{\sqrt{x/n_2}+x/n_2\}
 \mid\cF_{n_1}
 \right)\le2e^{-x}.
\]
Set $x=c\ell_n$ and union bound over $2k_n$ tagged coordinates.  Integration over Stage 1 proves the result.
\end{proof}

\begin{proof}[Proof of \Cref{thm:density}]
By \Cref{thm:preliminary},
$\max_j|q_j^0-Q_{j,n}|=O_{\Pp}(r_{1n})=o_{\Pp}(\delta_0)$, while
$h_n=o(\delta_0)$.  Hence, with probability tending to one,
$q_j^0\pm h_n$ lie in $\cN_n$ and in the interior of $[L,U]$ for every
$j$; endpoint projection is inactive.

Define the conditional population finite difference
\[
 f_{j,h}^0=\frac{F(q_j^0+h_n)-F(q_j^0-h_n)}{2h_n}
 =\frac1{2h_n}\int_{-h_n}^{h_n}f(q_j^0+u)\,du.
\]
For every $u\in[-h_n,h_n]$, \Cref{ass:smooth} gives
\begin{align*}
 |f(q_j^0+u)-f(Q_{j,n})|
 &\le L_f|q_j^0-Q_{j,n}+u|^\beta\\
 &\le L_f\{|q_j^0-Q_{j,n}|^\beta+|u|^\beta\},
\end{align*}
where $(x+y)^\beta\le x^\beta+y^\beta$ for $\beta\le1$.  Averaging over $u$ yields
\begin{equation}\label{eq:density-bias}
 \max_j|f_{j,h}^0-f(Q_{j,n})|
 \le C\left\{
 \max_{\ell\le k_n}|q_\ell^0-Q_{\ell,n}|^\beta+h_n^\beta
 \right\}
 =O_{\Pp}\{r_{1n}^\beta+h_n^\beta\}.
\end{equation}

By \Cref{lem:stage2-concentration},
\begin{align}
 \max_j|\wt f_j-f_{j,h}^0|
 &\le\frac1{2h_n}
 \max_j\left(
 |\ol G_{j,+}^{(2)}-F(q_j^0+h_n)|
 +|\ol G_{j,-}^{(2)}-F(q_j^0-h_n)|
 \right)\notag\\
 &\le C\frac{B_{2n}}{h_n}
 \left\{\sqrt{\frac{\ell_n}{n_2}}+\frac{\ell_n}{n_2}\right\}.
 \label{eq:density-stochastic}
\end{align}
Combining \eqref{eq:density-bias} and \eqref{eq:density-stochastic} gives the rate $d_n$ for $\wt f_j$.

Because $d_n\to0$, $f(Q_{j,n})\in[f_-,f_+]$, $\varpi_n\to0$, and $\varpi_n^{-1}\to\infty$, eventually
\[
 \varpi_n<f_-/2,
 \qquad
 \varpi_n^{-1}>2f_+,
\]
and the event $\max_j|\wt f_j-f(Q_{j,n})|<f_-/2$ has probability
tending to one.  On that event,
\[
 \frac{f_-}{2}
 \le \wt f_j
 \le f_++\frac{f_-}{2}
 \le\frac{3f_+}{2}
 <2f_+
 <\varpi_n^{-1},
\]
while $\varpi_n<f_-/2$.  Hence
$\varpi_n<\wt f_j<\varpi_n^{-1}$ for every $j$, so neither clipping
boundary is active and $\wh f_j=\wt f_j$.  This proves
\eqref{eq:density-rate}.
\end{proof}

\begin{proof}[Proof of \Cref{thm:onestep}]
Conditional on the first two stages, write
\[
 \ol G_j^{(3)}=F(q_j^0)+\ol Y_j,
 \qquad
 \ol Y_j=\frac1{n_3}\sum_{i=1}^{n_3}Y_{i,j}.
\]
Substituting this identity into \eqref{eq:onestep} and multiplying by $\wh f_j$ gives the exact equality
\begin{align}
 \wh f_j(\check Q_j-Q_{j,n})+\ol Y_j
 &=\wh f_j(q_j^0-Q_{j,n})-
   \{F(q_j^0)-F(Q_{j,n})\}.
 \label{eq:onestep-exact}
\end{align}
Add and subtract $f(Q_{j,n})(q_j^0-Q_{j,n})$ on the right:
\begin{align}
 \eqref{eq:onestep-exact}
 ={}&\{\wh f_j-f(Q_{j,n})\}(q_j^0-Q_{j,n})\notag\\
 &-\left[F(q_j^0)-F(Q_{j,n})
 -f(Q_{j,n})(q_j^0-Q_{j,n})\right].
 \label{eq:onestep-two-remainders}
\end{align}
The first term is uniformly $O_{\Pp}(d_nr_{1n})$ by \Cref{thm:preliminary,thm:density}.  For the second, the integral argument used in \eqref{eq:taylor-holder} gives
\[
 \left|F(q_j^0)-F(Q_{j,n})
 -f(Q_{j,n})(q_j^0-Q_{j,n})\right|
 \le\frac{L_f}{1+\beta}|q_j^0-Q_{j,n}|^{1+\beta},
\]
which is uniformly $O_{\Pp}(r_{1n}^{1+\beta})$.  Taking the maximum proves \eqref{eq:onestep-representation}.
\end{proof}

\section{High-Dimensional Gaussian and Multiplier Approximation}\label{app:bootstrap}

\begin{lemma}[Conditional variance and standardized envelope]\label{lem:variance-bounds}
Suppose \Cref{ass:iid,ass:smooth} hold and
\[
 \max_{j\le k_n}|q_j^0-Q_{j,n}|=o_{\Pp}(1).
\]
Then there exists a deterministic sequence $\eta_n\downarrow0$ such that,
with
\[
 \cE_{0,n}
 =\left\{\max_{j\le k_n}|q_j^0-Q_{j,n}|\le\eta_n\right\},
 \qquad
 \Pp(\cE_{0,n})\longrightarrow1,
\]
there are constants $0<c<C<\infty$, depending only on
$(\tau_-,\tau_+)$, for which the following bounds hold on $\cE_{0,n}$
for all sufficiently large $n$:
\[
 cH_{k_n}(1+\nu_{\eps_n})
 \le\min_{j\le k_n}\sigma_{j,n}^2
 \le\max_{j\le k_n}\sigma_{j,n}^2
 \le CH_{k_n}^2(1+\nu_{\eps_n}).
\]
Consequently, on the same event,
\begin{equation}\label{eq:K-bound}
 K_n\le C\sqrt{H_{k_n}}.
\end{equation}
\end{lemma}

\begin{proof}
Choose $\eta_n<\delta_0$ eventually.  On $\cE_{0,n}$, the segment joining
$Q_{j,n}$ and $q_j^0$ lies in $\cN_n$, and hence
\[
 |p_j^0-\tau_{j,n}|
 =|F(q_j^0)-F(Q_{j,n})|
 \le f_+\eta_n.
\]
Using the target-range part of \Cref{ass:iid}, for all sufficiently large
$n$,
\[
 p_j^0=F(q_j^0)\in[\tau_-/2,(1+\tau_+)/2]
\]
uniformly in $j$.  Thus both $p_j^0$ and $1-p_j^0$ are bounded below by
a positive constant.

For a threshold rank $r$, define $a=a_r^-(p_j^0)$ and $b=a_r^+(p_j^0)$.  Since $1\le w_r^\pm\le H_{k_n}$ and $2H_{k_n}-1\ge H_{k_n}$,
\[
 a\ge H_{k_n}\nu_{\eps_n}+cH_{k_n},
 \qquad
 b\ge H_{k_n}\nu_{\eps_n}+cH_{k_n}.
\]
The minimum over all fusion weights equals
\[
 p_j^0(1-p_j^0)+\frac{ab}{a+b}
 \ge c+\frac12\min(a,b)
 \ge cH_{k_n}(1+\nu_{\eps_n}).
\]
The variance at the actual anchor $\tau_j$ is no smaller than this minimum, proving the lower bound.

For the upper bound, $a$ and $b$ are at most
$CH_{k_n}^2(1+\nu_{\eps_n})$, and \eqref{eq:quadvar} at any $\lambda\in[0,1]$ is bounded by the same order.  Alternatively, the score envelope gives a coarser bound of the same required form.  Finally,
\[
 1+\nu_\eps
 =1+\frac{4e^\eps}{(e^\eps-1)^2}
 =\left(\frac{e^\eps+1}{e^\eps-1}\right)^2,
\]
so
\[
 B_{1n}
 =1+\frac{2H_{k_n}e^{\eps_n}}{e^{\eps_n}-1}
 \le3H_{k_n}\sqrt{1+\nu_{\eps_n}}.
\]
Therefore
\[
 \frac{|Y_{i,j}|}{\sigma_{j,n}}
 \le\frac{2B_{1n}}
 {\{cH_{k_n}(1+\nu_{\eps_n})\}^{1/2}}
 \le C\sqrt{H_{k_n}},
\]
which proves \eqref{eq:K-bound}.
\end{proof}

\begin{lemma}[Uniform sample-scale consistency]\label{lem:sample-scale}
Conditional on the first two stages,
\begin{equation}\label{eq:sample-scale-rate}
 \max_{j\le k_n}
 \left|\frac{\wh\sigma_{G,j}^2}{\sigma_{j,n}^2}-1\right|
 =O_{\Pp}\left[
 K_n\sqrt{\frac{\ell_n}{n_3}}
 +K_n^2\frac{\ell_n}{n_3}
 \right].
\end{equation}
Under \eqref{eq:hdclt-cond}, the right-hand side is $o_{\Pp}(1)$.
\end{lemma}

\begin{proof}
Let $X_{i,j}=Y_{i,j}/\sigma_{j,n}$ and $X_i=(X_{i,1},\ldots,X_{i,k_n})^\top$.  Conditional on the first two stages, the vectors are i.i.d., $\E X_{i,j}=0$, $\E X_{i,j}^2=1$, and $|X_{i,j}|\le K_n$.  In particular, $\E X_{i,j}^4\le K_n^2$.  Bernstein's inequality and a union bound give
\[
 \max_j\left|n_3^{-1}\sum_iX_{i,j}^2-1\right|
 =O_{\Pp}\left\{K_n\sqrt{\ell_n/n_3}
 +K_n^2\ell_n/n_3\right\}.
\]
The squared sample means satisfy
\[
 \max_j\left|n_3^{-1}\sum_iX_{i,j}\right|^2
 =O_{\Pp}(K_n^2\ell_n/n_3),
\]
again by Bernstein.  Since
\[
 \frac{\wh\sigma_{G,j}^2}{\sigma_{j,n}^2}
 =n_3^{-1}\sum_iX_{i,j}^2
 -\left(n_3^{-1}\sum_iX_{i,j}\right)^2,
\]
the result follows.
\end{proof}

\begin{lemma}[Conditional Gaussian and empirical multiplier approximation]\label{lem:cck}
Let
\[
 S_n^\dagger=\max_{j\in\widehat{\mathcal J}_n}
 \left|\frac{n_3^{-1/2}\sum_{i=1}^{n_3}Y_{i,j}}
 {\wh\sigma_{G,j}}\right|,
 \qquad
 \Delta_{\sigma,n}=K_n\sqrt{\frac{\ell_n}{n_3}}
 +K_n^2\frac{\ell_n}{n_3},
\]
and let $S_n^*$ be the empirical Gaussian multiplier analogue centered by the sample means and divided by the sample standard deviations.  Under \eqref{eq:hdclt-cond}, conditional on the first two stages,
\begin{equation}\label{eq:cck-bound}
 \begin{split}
 &\sup_z\left|
 \Pp(S_n^\dagger\le z\mid\cF_{n_1+n_2})
 -\Pp(S_n^*\le z\mid G_1^{(3)},\ldots,G_{n_3}^{(3)})
 \right|\\
 &\quad=O_{\Pp}\left\{
 \left(\frac{K_n^2\ell_n^7}{n_3}\right)^{1/6}
 +\ell_n\Delta_{\sigma,n}\right\}=o_{\Pp}(1).
 \end{split}
\end{equation}
\end{lemma}

\begin{proof}
Work conditionally on $\cF_{n_1+n_2}$.  Put
\[
 X_{i,j}=\frac{Y_{i,j}}{\sigma_{j,n}},
 \qquad
 p_n=\max\{2k_n,3\}.
\]
Define the population-standardized target and multiplier maxima
\begin{align*}
 S_n^0
 &=\max_{j\le k_n}
 \left|n_3^{-1/2}\sum_{i=1}^{n_3}X_{i,j}\right|,\\
 S_n^{*,0}
 &=\max_{j\le k_n}
 \left|n_3^{-1/2}\sum_{i=1}^{n_3}
 \xi_i\left(X_{i,j}-n_3^{-1}\sum_{\ell=1}^{n_3}X_{\ell,j}\right)\right|.
\end{align*}
Let $Z_n$ be the maximum of the absolute coordinates of the centered
Gaussian vector with the covariance matrix of
$n_3^{-1/2}\sum_iX_i$, and let $F_n^G$ be its conditional CDF.

Embed the absolute maximum into $(X_i^\top,-X_i^\top)^\top$ and, if
$2k_n<3$, append a duplicate coordinate.  Every coordinate is centered,
has variance one, and is bounded by $K_n$.  Furthermore,
\[
 \log(p_nn_3)\le C\ell_n,
\]
and, when $k_n\ge2$, $\log(p_nn_3)\le\ell_n$.  For $m=1,2$,
\[
 \E(|X_{i,j}|^{2+m}\mid\cF_{n_1+n_2})
 \le K_n^m\E(X_{i,j}^2\mid\cF_{n_1+n_2})
 =K_n^m.
\]
Also, after replacing $K_n$ by a universal multiple,
\[
 \E\{\exp(|X_{i,j}|/(CK_n))\mid\cF_{n_1+n_2}\}\le2.
\]
Hence Proposition~2.1 of \citet{CCK2017} gives
\begin{equation}\label{eq:target-gaussian-raw}
 \sup_z\left|
 \Pp(S_n^0\le z\mid\cF_{n_1+n_2})-F_n^G(z)
 \right|
 \le C\left(\frac{K_n^2\ell_n^7}{n_3}\right)^{1/6}.
\end{equation}

Put $m=n_3$ and $\rho_n=K_n^2\ell_n^7/m$.  For the multiplier
comparison, apply Corollary~4.2 of \citet{CCK2017} with
$\eta=(m+1)^{-2}$.  Conditional on the first two stages, with probability
at least $1-(m+1)^{-2}$ over the Stage-3 sample,
\begin{equation}\label{eq:bootstrap-gaussian-raw}
 \sup_z\left|
 \Pp(S_n^{*,0}\le z\mid X_1,\ldots,X_m)-F_n^G(z)
 \right|
 \le C\rho_n^{1/6},
\end{equation}
because
$\log^5(p_nm)\log^2\{(m+1)^2\}\le C\ell_n^7$.

It remains to replace the population scales by sample scales.  Write
\[
 \delta_{\sigma,n}=\max_j
 \left|\frac{\wh\sigma_{G,j}^2}{\sigma_{j,n}^2}-1\right|,
 \qquad
 \ol X_j=m^{-1}\sum_{i=1}^mX_{i,j}.
\]
For each coordinate, $|X_{i,j}|\le K_n$,
$\E(X_{i,j}^2\mid\cF_{n_1+n_2})=1$, and
$\E(X_{i,j}^4\mid\cF_{n_1+n_2})\le K_n^2$.  Conditional Bernstein
inequalities and a union bound therefore give
\begin{align}
 \Pp\left[
 \max_j\left|m^{-1}\sum_{i=1}^m(X_{i,j}^2-1)\right|
 >C\Delta_{\sigma,n}
 \;\middle|\;\cF_{n_1+n_2}\right]
 &\le Ce^{-c\ell_n},\label{eq:scale-second-moment-tail}\\
 \Pp\left[
 \max_j|\ol X_j|>
 C\left\{\sqrt{\frac{\ell_n}{m}}+
 K_n\frac{\ell_n}{m}\right\}
 \;\middle|\;\cF_{n_1+n_2}\right]
 &\le Ce^{-c\ell_n}.\label{eq:scale-mean-tail}
\end{align}
The constants in the thresholds can be enlarged so that $c$ is
arbitrarily large.  On first-two-stage histories with $\rho_n$ smaller
than a fixed constant, the square of the threshold in
\eqref{eq:scale-mean-tail} is bounded by $C\Delta_{\sigma,n}$.
Since
\[
 \frac{\wh\sigma_{G,j}^2}{\sigma_{j,n}^2}
 =m^{-1}\sum_{i=1}^mX_{i,j}^2-\ol X_j^2,
\]
\eqref{eq:scale-second-moment-tail}--\eqref{eq:scale-mean-tail} imply
\begin{equation}\label{eq:scale-explicit-tail}
 \Pp(\delta_{\sigma,n}>C\Delta_{\sigma,n}
 \mid\cF_{n_1+n_2})\le Ce^{-c\ell_n}.
\end{equation}
Moreover, $\rho_n=o_{\Pp}(1)$ by \eqref{eq:hdclt-cond}, and on these
histories
\[
 \Delta_{\sigma,n}
 =\rho_n^{1/2}\ell_n^{-3}+\rho_n\ell_n^{-6}=o(1).
\]
Thus \eqref{eq:scale-explicit-tail} also ensures
$\delta_{\sigma,n}\le1/2$ with the stated probability.

On $\{\delta_{\sigma,n}\le1/2\}$ all sample scales are positive and
\begin{align*}
 |S_n^\dagger-S_n^0|
 &\le C\delta_{\sigma,n}S_n^0,\\
 |S_n^*-S_n^{*,0}|
 &\le C\delta_{\sigma,n}S_n^{*,0}.
\end{align*}
Another conditional Bernstein inequality and union bound give
\[
 \Pp(S_n^0>C\sqrt{\ell_n}\mid\cF_{n_1+n_2})
 \le Ce^{-c\ell_n}.
\]
On the sample-scale event, every coordinate of the conditional Gaussian
vector defining $S_n^{*,0}$ has variance at most $3/2$, so a Gaussian
union bound gives
\[
 \Pp(S_n^{*,0}>C\sqrt{\ell_n}\mid X_1,\ldots,X_m)
 \le Ce^{-c\ell_n}.
\]
Consequently, with $t_n=C\Delta_{\sigma,n}\sqrt{\ell_n}$, the two
studentization perturbations exceed $t_n$ only on the explicit
exceptional events above.  Nazarov's inequality for the signed Gaussian
vector gives
\[
 \sup_z\{F_n^G(z+t_n)-F_n^G(z-t_n)\}
 \le C\ell_n\Delta_{\sigma,n}.
\]
The exponentially small terms can be absorbed into $C\rho_n^{1/6}$:
$K_n\ge1$, $m\le n$, and the exponent $c$ can be chosen sufficiently
large.  Combining the perturbation sandwiches separately with
\eqref{eq:target-gaussian-raw} and
\eqref{eq:bootstrap-gaussian-raw} gives
\begin{align}
 \sup_z\left|
 \Pp(S_n^\dagger\le z\mid\cF_{n_1+n_2})-F_n^G(z)
 \right|
 &=O_{\Pp}\left\{
 \left(\frac{K_n^2\ell_n^7}{n_3}\right)^{1/6}
 +\ell_n\Delta_{\sigma,n}\right\},
 \label{eq:target-gaussian-cdf}\\
 \sup_z\left|
 \Pp(S_n^*\le z\mid G_1^{(3)},\ldots,G_{n_3}^{(3)})-F_n^G(z)
 \right|
 &=O_{\Pp}\left\{
 \left(\frac{K_n^2\ell_n^7}{n_3}\right)^{1/6}
 +\ell_n\Delta_{\sigma,n}\right\}
 \label{eq:bootstrap-gaussian-cdf}
\end{align}
on a set of first-two-stage histories whose probability tends to one; the
second display also holds outside a Stage-3 event of conditional
probability at most $C e^{-c\ell_n}+(m+1)^{-2}$.  The triangle inequality
then proves \eqref{eq:cck-bound}.

Finally,
\[
 \ell_n\Delta_{\sigma,n}
 =\left(\frac{K_n^2\ell_n^7}{n_3}\right)^{1/2}\ell_n^{-2}
 +\frac{K_n^2\ell_n^7}{n_3}\ell_n^{-5}
 =o_{\Pp}(1),
\]
which completes the proof.
\end{proof}

\begin{lemma}[Transfer of the one-step remainder]\label{lem:remainder-transfer}
Suppose \eqref{eq:onestep-representation},
\eqref{eq:sample-scale-rate}, $\Delta_{\sigma,n}=o_{\Pp}(1)$, and
\eqref{eq:remainder-cond} hold.  Then
\begin{equation}\label{eq:T-distance}
 |T_n-S_n^\dagger|
 =O_{\Pp}\left(
 \frac{\sqrt{n_3}R_n}{\ul\sigma_n}
 \right)
 =o_{\Pp}(\ell_n^{-1/2}).
\end{equation}
\end{lemma}

\begin{proof}
By \Cref{thm:onestep}, uniformly in $j$,
\[
 \left|\sqrt{n_3}\wh f_j(\check Q_j-Q_{j,n})
 +n_3^{-1/2}\sum_iY_{i,j}\right|
 =O_{\Pp}(\sqrt{n_3}R_n).
\]
On the event
\[
 \max_j\left|\frac{\wh\sigma_{G,j}^2}{\sigma_{j,n}^2}-1\right|
 \le\frac12,
\]
which has probability tending to one by \Cref{lem:sample-scale},
$\widehat{\mathcal J}_n=\{1,\ldots,k_n\}$ and
$\min_j\wh\sigma_{G,j}\ge\ul\sigma_n/\sqrt2$.  Divide by the sample
scales, maximize, and use the reverse triangle inequality for maxima.
Condition \eqref{eq:remainder-cond} then proves \eqref{eq:T-distance}.
\end{proof}

\begin{proof}[Proof of \Cref{thm:bootstrap}]
Condition on $\cF_{n_1+n_2}$.  Let $Z_n$ be the maximum of the absolute
coordinates ${Z_{n,j}}$ of the centered Gaussian vector used in \Cref{lem:cck}, let
$F_n^G$ be its conditional CDF, and put
\[
 q_n^G(u)=\inf\{z:F_n^G(z)\ge u\}.
\]
The two comparisons in \eqref{eq:target-gaussian-cdf} and
\eqref{eq:bootstrap-gaussian-cdf}, together with
$\Pp(\widehat{\mathcal J}_n=\{1,\ldots,k_n\})\to1$, imply that there are a
deterministic sequence ${\delta_{G,n}}\downarrow0$ and events $\mathcal A_n$
with $\Pp(\mathcal A_n^c)=o(1)$ such that, on $\mathcal A_n$,
\begin{align}
 \sup_z\left|
 \Pp(S_n^\dagger\le z\mid\cF_{n_1+n_2})-F_n^G(z)
 \right|&\le{\delta_{G,n}},\label{eq:target-comparison-event}\\
 \sup_z\left|
 \Pp(S_n^*\le z\mid G_1^{(3)},\ldots,G_{n_3}^{(3)})-F_n^G(z)
 \right|&\le{\delta_{G,n}},\label{eq:bootstrap-comparison-event}
\end{align}
and all sample variances are positive.  This event formulation follows
directly from the two $O_{\Pp}$ bounds because their common rate tends to
zero.

Every coordinate of the Gaussian vector has variance one.  Consequently,
for every $z$,
\[
 \Pp(Z_n=z\mid\cF_{n_1+n_2})
 \le\sum_{j=1}^{k_n}\Pp(|Z_{n,j}|=z\mid\cF_{n_1+n_2})=0.
\]
Thus $F_n^G$ is continuous even if its correlation matrix is singular, and
$F_n^G\{q_n^G(u)\}=u$ for $u\in(0,1)$.  Write $u=1-\alpha$ and $c_u^*=c_{1-\alpha}^*$.  The second
inequality in \eqref{eq:bootstrap-comparison-event} and the definition of a
generalized quantile give, on $\mathcal A_n$,
\begin{equation}\label{eq:exact-critical-sandwich}
 q_n^G(u-{\delta_{G,n}})
 \le c_u^*
 \le q_n^G(u+{\delta_{G,n}}).
\end{equation}
Indeed, at a point below the left endpoint the conditional multiplier CDF
is strictly below $u$, whereas at the right endpoint it is at least $u$.

The sandwich \eqref{eq:exact-critical-sandwich} implies the pathwise
inclusions
\[
 \mathcal A_n\cap\{S_n^\dagger\le q_n^G(u-{\delta_{G,n}})\}
 \subseteq\{S_n^\dagger\le c_u^*\}
 \subseteq
 \mathcal A_n^c\cup\{S_n^\dagger\le q_n^G(u+{\delta_{G,n}})\}.
\]
Using \eqref{eq:target-comparison-event} at the two deterministic,
$\cF_{n_1+n_2}$-measurable Gaussian quantiles and then removing
$\mathcal A_n$ yields, uniformly for
$u\in[\alpha_0,1-\alpha_0]$,
\[
 \left|\Pp(S_n^\dagger\le c_u^*)-u\right|
 \le 2{\delta_{G,n}}+\Pp(\mathcal A_n^c)=o(1).
\]

By \Cref{lem:remainder-transfer}, choose a deterministic
$a_n=o(\ell_n^{-1/2})$ such that, for
$\mathcal D_n=\{|T_n-S_n^\dagger|\le a_n\}$,
$\Pp(\mathcal D_n^c)=o(1)$.  On $\mathcal A_n\cap\mathcal D_n$,
\begin{align*}
 \{S_n^\dagger\le q_n^G(u-{\delta_{G,n}})-a_n\}
 &\subseteq\{T_n\le c_u^*\},\\
 \{T_n\le c_u^*\}
 &\subseteq\{S_n^\dagger\le q_n^G(u+{\delta_{G,n}})+a_n\}.
\end{align*}
Nazarov's inequality, applied to the $2k_n$-dimensional signed Gaussian
vector, gives
\[
 \sup_z\{F_n^G(z+a_n)-F_n^G(z-a_n)\}
 \le Ca_n\sqrt{\ell_n}=o(1).
\]
Combining this bound with \eqref{eq:target-comparison-event} proves
\eqref{eq:bootstrap-validity}.  On the event that all sample variances are
positive, $T_n\le c_{1-\alpha}^*$ is algebraically equivalent to
simultaneous inclusion in the untruncated intervals $\cI_{n,j}^*$.
Intersecting with $[L,U]$ preserves inclusion because
$Q_{j,n}\in[L,U]$.  The zero-variance fallback is $[L,U]$ and its
probability tends to zero, which proves \eqref{eq:coverage}.

For the Monte Carlo version, let $F_n^*$ be the exact conditional
multiplier CDF and $\widehat F_{n,M}^*$ its empirical CDF based on $M_n$
independent draws.  Also write
\[
 q_n^*(v)=\inf\{z:F_n^*(z)\ge v\}.
\]
Conditionally on the data, the Dvoretzky--Kiefer--Wolfowitz (DKW) inequality \citep{Massart1990} gives
\begin{equation}\label{eq:dkw-finite-m}
 \Pp\left\{
 \sup_z|\widehat F_{n,M}^*(z)-F_n^*(z)|>\eta
 \mathrel{\Big|}G_1^{(3)},\ldots,G_{n_3}^{(3)}
 \right\}\le2e^{-2M_n\eta^2}.
\end{equation}
For $u\in[\alpha_0,1-\alpha_0]$ and all sufficiently large $M_n$, the cap
in $r_{M_n,\alpha}$ is inactive.  At
$\widehat c_u^*=T_{(r_{M_n,\alpha})}^*$,
\[
 \widehat F_{n,M}^*(\widehat c_u^*)\ge u,
 \qquad
 \widehat F_{n,M}^*(\widehat c_u^*-)<u+M_n^{-1}.
\]
On the DKW event in \eqref{eq:dkw-finite-m}, these two inequalities imply
\begin{equation}\label{eq:finite-M-quantile-sandwich}
 q_n^*(u-\eta)
 \le\widehat c_u^*
 \le q_n^*(u+\eta+M_n^{-1}).
\end{equation}
For the lower bound, a value below $q_n^*(u-\eta)$ would have empirical CDF
strictly below $u$; for the upper bound, a value above
$q_n^*(u+\eta+M_n^{-1})$ would make the empirical left limit at the latter
quantile at least $u+M_n^{-1}$, contradicting the preceding display.
Take $\eta=M_n^{-1/4}$ in \eqref{eq:dkw-finite-m}.  Then the DKW failure
probability tends to zero, both shifted levels remain in a fixed compact
subset of $(0,1)$, and \eqref{eq:finite-M-quantile-sandwich} together with
the exact-quantile result proves the two finite-$M_n$ assertions uniformly
in $\alpha$.

Now fix $M$ and $\alpha$, and abbreviate $r=r_{M,\alpha}$.  On the event
of positive sample variances, $F_n^*$ is continuous because every
studentized multiplier coordinate is a nondegenerate Gaussian variable.
The preceding Gaussian comparisons and the remainder transfer imply
\[
 \sup_z\left|
 \Pp(T_n\le z\mid\cF_{n_1+n_2})-F_n^G(z)
 \right|=o_{\Pp}(1),
 \qquad
 \sup_z|F_n^*(z)-F_n^G(z)|=o_{\Pp}(1).
\]
Since $F_n^G$ is continuous, the conditional probability-integral
transform gives
\[
 U_n:=F_n^*(T_n)
 \rightsquigarrow U,
 \qquad U\sim\operatorname{Uniform}(0,1).
\]
Conditional on the data, $T_n\le T_{(r)}^*$ exactly when at most $r-1$
of the $M$ multiplier maxima are below $T_n$.  Therefore
\[
 \Pp(T_n\le T_{(r)}^*\mid G_1^{(3)},\ldots,G_{n_3}^{(3)})
 =g_{M,r}(U_n),
\]
where
\[
 g_{M,r}(u)=\sum_{s=0}^{r-1}{M\choose s}u^s(1-u)^{M-s}.
\]
This is a bounded continuous polynomial, and the beta integral gives
\[
 \int_0^1g_{M,r}(u)\,du
 =\sum_{s=0}^{r-1}{M\choose s}
 \frac{s!(M-s)!}{(M+1)!}
 =\frac{r}{M+1}.
\]
Taking expectations proves \eqref{eq:fixed-M-coverage}.  The interval
equivalence and the vanishing zero-variance fallback then give the same
limit for simultaneous coverage.
\end{proof}

\begin{proof}[Proof of \Cref{cor:width}]
Conditional on the Stage-3 data, every nondegenerate coordinate of the
studentized Gaussian multiplier vector has variance one.  Hence
\[
 \Pp(S_n^*>x\mid G_1^{(3)},\ldots,G_{n_3}^{(3)})
 \le2k_ne^{-x^2/2}.
\]
It follows that the exact conditional critical value satisfies
\[
 c_{1-\alpha}^*
 \le\sqrt{2\log(2k_n/\alpha)}
 =O\{\sqrt{\log(2k_n)}\}
\]
uniformly for $\alpha$ bounded away from zero.  For the empirical critical
value, \eqref{eq:finite-M-quantile-sandwich} and $M_n\to\infty$ give the
same order in probability.

By \Cref{lem:sample-scale,lem:variance-bounds},
\[
 \max_j\wh\sigma_{G,j}
 =O_{\Pp}\{H_{k_n}\sqrt{1+\nu_{\eps_n}}\}.
\]
By \Cref{thm:density},
$\min_j\wh f_j\ge f_-/2$ with probability tending to one.  Substitution in
\eqref{eq:intervals} proves \eqref{eq:width-general}.  For
$0<\eps_n\le1$,
\[
 \sqrt{1+\nu_{\eps_n}}
 =\frac{e^{\eps_n}+1}{e^{\eps_n}-1}
 \le\frac{C}{\eps_n},
\]
which proves \eqref{eq:width-small-eps}.

It remains to verify the transparent regime.  Write
\[
 H=H_{k_n}.
\]
Since
\[
 H_{2k_n}
 =\left\lceil\log_2(2k_n+1)\right\rceil
 \le1+\left\lceil\log_2(k_n+1)\right\rceil
 =H+1\le2H,
\]
and $\eps_n$ is bounded away from zero,
\[
 B_{1n}\asymp B_{2n}\asymp H.
\]
Moreover, $H\le C\ell_n$.  By \Cref{lem:variance-bounds},
\[
 \ul\sigma_n\gtrsim H^{1/2},
 \qquad K_n^2=O(H).
\]

Under $a=2/3$, $\beta=1$, and fixed positive stage fractions,
\eqref{eq:r1n} gives
\[
 r_{1n}
 =O\left(
 \frac{b_n}{n}
+H\sqrt{\frac{\ell_n}{n}}
+Hn^{-2/3}\sqrt{\ell_n}
+H^2n^{-2/3}\ell_n
 \right).
\]
Using $b_n/n=o(\ell_n^{3/2}n^{-1/2})$ and $H\le C\ell_n$,
\[
 r_{1n}
 =O\left(
 \ell_n^{3/2}n^{-1/2}
 +\ell_n^3n^{-2/3}
 \right).
\]
The ratio of the second term to the first is
\[
 \frac{\ell_n^3n^{-2/3}}{\ell_n^{3/2}n^{-1/2}}
 =\ell_n^{3/2}n^{-1/6}
 =\left(\frac{\ell_n^9}{n}\right)^{1/6}
 =o(1).
\]
Therefore,
\begin{equation}\label{eq:r1n-transparent-final}
 r_{1n}=O(\ell_n^{3/2}n^{-1/2}).
\end{equation}

The proposed bandwidth obeys
\[
 h_n
 \asymp
 \left(\frac{H^2\ell_n}{n}\right)^{1/4}
 =H^{1/2}\ell_n^{1/4}n^{-1/4}.
\]
Its leading stochastic density term is
\[
 \frac{B_{2n}}{h_n}\sqrt{\frac{\ell_n}{n}}
 \asymp h_n,
\]
whereas
\[
 \frac{B_{2n}}{h_n}\frac{\ell_n}{n}
 =O\left(h_n\sqrt{\frac{\ell_n}{n}}\right)
 =o(h_n).
\]
Also,
\[
 \frac{r_{1n}}{h_n}
 =O\left(\frac{\ell_n^{5/4}}{H^{1/2}n^{1/4}}\right)
 =o(1),
\]
because its fourth power is bounded by $C\ell_n^5/n=o(1)$.
Consequently,
\begin{equation}\label{eq:dn-transparent}
 d_n=O\left(H^{1/2}\ell_n^{1/4}n^{-1/4}\right).
\end{equation}

Combining \eqref{eq:r1n-transparent-final} and
\eqref{eq:dn-transparent} gives
\[
 r_{1n}d_n
 =O\left(H^{1/2}\ell_n^{7/4}n^{-3/4}\right).
\]
Since $d_n\ge h_n$ and $r_{1n}=o(h_n)$, we have
$r_{1n}=o(d_n)$ and hence $r_{1n}^2=o(r_{1n}d_n)$.  Therefore
\[
 R_n
 =O\left(H^{1/2}\ell_n^{7/4}n^{-3/4}\right).
\]
It follows that
\[
 \frac{\sqrt{n_3\ell_n}}{\ul\sigma_n}R_n
 =O(\ell_n^{9/4}n^{-1/4})
 =O\left\{\left(\frac{\ell_n^9}{n}\right)^{1/4}\right\}
 =o(1).
\]
The Gaussian approximation condition is
\[
 \frac{K_n^2\ell_n^7}{n_3}
 =O\left(\frac{H\ell_n^7}{n}\right)
 =O\left(\frac{\ell_n^8}{n}\right)
 =o(1).
\]

Finally, choose $\omega_n=\ell_n$ in the primitive construction of
\Cref{lem:localization}.  Then
\[
 B_{1n}^2\ell_n\omega_n
 =O(H^2\ell_n^2)
 \le O(\ell_n^4)
 =o(n^{2/3}),
\]
and
\begin{align*}
 b_n
 &=O\left\{(H^2\ell_n^2)^{3/2}+(\log n)^3\right\}\\
 &=O\{H^3\ell_n^3+(\log n)^3\}
 =O(\ell_n^6).
\end{align*}
Moreover,
\[
 \frac{b_n/n}{\ell_n^{3/2}n^{-1/2}}
 =O\left(\frac{\ell_n^{9/2}}{\sqrt n}\right)
 =O\left\{\left(\frac{\ell_n^9}{n}\right)^{1/2}\right\}
 =o(1).
\]
Thus the primitive construction satisfies all three localization conditions
and the additional transient requirement in the corollary.

Because $\ell_n=\log\{2k_n(n+1)\}$, condition
\eqref{eq:transparent-growth} is equivalent to
\[
 \log k_n=o(n^{1/9}).
\]
It therefore includes fixed grids, arbitrarily slowly growing grids, all
polynomial grids, and more generally $k_n=\exp\{o(n^{1/9})\}$.
\end{proof}

\section{Proofs for Post-Processing and Fixed-Chain Comparisons}\label{app:comparisons}

\begin{proof}[Proof of \Cref{prop:monotone-refinement}]
Fix $j$.  On the original simultaneous coverage event, for every $\ell\le j$,
\[
 L_\ell^I\le Q_\ell\le Q_j,
\]
where the second inequality follows from monotonicity of the quantile function.  Therefore $\wt L_j^I=\max_{\ell\le j}L_\ell^I\le Q_j$.  Similarly, for every $\ell\ge j$,
\[
 Q_j\le Q_\ell\le U_\ell^I,
\]
so $Q_j\le\min_{\ell\ge j}U_\ell^I=\wt U_j^I$.  Both inequalities hold for every $j$ on the same event.
\end{proof}

\begin{proof}[Proof of \Cref{prop:fixedJ}]
Let
\[
 s_J^2=(J-1)^{-1}\sum_{b=1}^J(W_b-\ol W)^2.
\]
For a normal sample, $\sqrt J(\ol W-\theta)/s_J\sim t_{J-1}$.  Because
\[
 S_J^2=\frac{J-1}{J}s_J^2,
\]
we have
\[
 \frac{\sqrt{J-1}(\ol W-\theta)}{S_J}
 =\frac{\sqrt J(\ol W-\theta)}{s_J}
 \sim t_{J-1}.
\]
Conditional on $W_1,\ldots,W_J$, the multiplier numerator is centered normal with variance
\[
 \Var\left(J^{-1/2}\sum_{b=1}^J\xi_b(W_b-\ol W)\mid W_1,\ldots,W_J\right)
 =J^{-1}\sum_{b=1}^J(W_b-\ol W)^2
 =S_J^2.
\]
Division by $S_J$ therefore gives an exact conditional $N(0,1)$ variable.  Neither distribution changes when an external within-block sample size increases while $J$ remains fixed.
\end{proof}

\begin{proof}[Proof of \Cref{prop:bonferroni}]
The upper tail of a $t_\nu$ variable is regularly varying with index $-\nu$:
\[
 \Pp(t_\nu>x)\sim C_\nu x^{-\nu}
 \qquad (x\to\infty)
\]
for a positive constant $C_\nu$.  Setting the tail probability equal to $\alpha/(2k)$ and inverting gives
\[
 t_{\nu,1-\alpha/(2k)}
 \sim\left(\frac{2C_\nu k}{\alpha}\right)^{1/\nu},
\]
which proves \eqref{eq:t-growth}.  The most stringent Holm step uses the same order of tail probability.  For a standardized Gaussian vector $(Z_1,\ldots,Z_k)$, regardless of dependence,
\[
 \Pp\left(\max_{j\le k}|Z_j|>x\right)
 \le2k e^{-x^2/2}.
\]
Taking $x=\sqrt{2\log(2k/\alpha)}$ proves the $O(\sqrt{\log k})$ upper order for a Gaussian critical value.
\end{proof}

\begin{thebibliography}{99}

\bibitem[Aamand et~al.(2025)Aamand, Boninsegna, Gentle, Imola, and Pagh]{Aamand2025}
Aamand, A., Boninsegna, F., Gentle, A., Imola, J., and Pagh, R. (2025).
Lightweight protocols for distributed private quantile estimation.
\emph{Proceedings of the 42nd International Conference on Machine Learning}, PMLR 267, 27--58.

\bibitem[Acharya et~al.(2019)Acharya, Sun, and Zhang]{Acharya2019}
Acharya, J., Sun, Z., and Zhang, H. (2019).
Hadamard response: Estimating distributions privately, efficiently, and with little communication.
\emph{Proceedings of the 22nd International Conference on Artificial Intelligence and Statistics}, PMLR 89, 1120--1129.

\bibitem[Anderson(2003)]{Anderson2003}
Anderson, T. W. (2003).
\emph{An Introduction to Multivariate Statistical Analysis}, 3rd ed.
Wiley, Hoboken, NJ.

\bibitem[Autor et~al.(2006)Autor, Katz, and Kearney]{AutorKatzKearney2006}
Autor, D. H., Katz, L. F., and Kearney, M. S. (2006).
The polarization of the U.S. labor market.
\emph{American Economic Review} \textbf{96}, 189--194.

\bibitem[Bahadur(1966)]{Bahadur1966}
Bahadur, R. R. (1966).
A note on quantiles in large samples.
\emph{Annals of Mathematical Statistics} \textbf{37}, 577--580.

\bibitem[Cai et~al.(2025)Cai, Hu, Sun, and Wu]{Cai2025}
Cai, L., Hu, Q., Sun, J., and Wu, S. (2025).
Time-uniform and asymptotic confidence sequence of quantile under local differential privacy.
\emph{Advances in Neural Information Processing Systems} \textbf{38}, 114488--114520.

\bibitem[Cai et~al.(2026a)Cai, Hu, and Wu]{Cai2026Fed}
Cai, L., Hu, Q., and Wu, S. (2026a).
Federated learning of quantile inference under local differential privacy.
\emph{The Fourteenth International Conference on Learning Representations}.

\bibitem[Cai et~al.(2026b)Cai, Hu, and Wu]{Cai2026Integration}
Cai, L., Hu, Q., and Wu, S. (2026b).
Privacy-aware data integration for enhanced quantile inference under heterogeneity.
\emph{Proceedings of the 43rd International Conference on Machine Learning}, PMLR 306, 10556--10590.

\bibitem[Chernozhukov et~al.(2013a)Chernozhukov, Chetverikov, and Kato]{CCK2013}
Chernozhukov, V., Chetverikov, D., and Kato, K. (2013a).
Gaussian approximations and multiplier bootstrap for maxima of sums of high-dimensional random vectors.
\emph{Annals of Statistics} \textbf{41}, 2786--2819.

\bibitem[Chernozhukov et~al.(2017)Chernozhukov, Chetverikov, and Kato]{CCK2017}
Chernozhukov, V., Chetverikov, D., and Kato, K. (2017).
Central limit theorems and bootstrap in high dimensions.
\emph{Annals of Probability} \textbf{45}, 2309--2352.

\begingroup
\bibitem[Chernozhukov et~al.(2013b)Chernozhukov, Fern\'andez-Val, and Melly]{CFM2013}
Chernozhukov, V., Fern\'andez-Val, I., and Melly, B. (2013b).
Inference on counterfactual distributions.
\emph{Econometrica} \textbf{81}, 2205--2268.
\endgroup

\bibitem[Cormode et~al.(2019)Cormode, Kulkarni, and Srivastava]{Cormode2019}
Cormode, G., Kulkarni, T., and Srivastava, D. (2019).
Answering range queries under local differential privacy.
\emph{Proceedings of the VLDB Endowment} \textbf{12}, 1126--1138.

\begingroup
\bibitem[Cs\"org\H{o}(1983)]{Csorgo1983}
Cs\"org\H{o}, M. (1983).
\emph{Quantile Processes with Statistical Applications}.
CBMS-NSF Regional Conference Series in Applied Mathematics 42, SIAM, Philadelphia.
\endgroup

\bibitem[Dette and Graw(2024)]{DetteGraw2024}
Dette, H. and Graw, C. (2024).
Uncertainty quantification by block bootstrap for differentially private stochastic gradient descent.
arXiv:2405.12553.

\bibitem[Duchi et~al.(2018)Duchi, Jordan, and Wainwright]{Duchi2018}
Duchi, J. C., Jordan, M. I., and Wainwright, M. J. (2018).
Minimax optimal procedures for locally private estimation.
\emph{Journal of the American Statistical Association} \textbf{113}, 182--201.

\bibitem[Holm(1979)]{Holm1979}
Holm, S. (1979).
A simple sequentially rejective multiple test procedure.
\emph{Scandinavian Journal of Statistics} \textbf{6}, 65--70.

\bibitem[Kairouz et~al.(2014)Kairouz, Oh, and Viswanath]{Kairouz2014}
Kairouz, P., Oh, S., and Viswanath, P. (2014).
Extremal mechanisms for local differential privacy.
\emph{Advances in Neural Information Processing Systems} \textbf{27}.

\bibitem[Kasiviswanathan et~al.(2011)Kasiviswanathan, Lee, Nissim, Raskhodnikova, and Smith]{Kasiviswanathan2011}
Kasiviswanathan, S. P., Lee, H. K., Nissim, K., Raskhodnikova, S., and Smith, A. (2011).
What can we learn privately?
\emph{SIAM Journal on Computing} \textbf{40}, 793--826.

\bibitem[Kushner and Yin(2003)]{KushnerYin2003}
Kushner, H. J. and Yin, G. G. (2003).
\emph{Stochastic Approximation and Recursive Algorithms and Applications}, 2nd ed.
Springer, New York.

\bibitem[Liu and Hu(2026)]{LiuHu2026QR}
Liu, Y. and Hu, Q. (2026).
Locally private online quantile regression: Estimation and inference.
arXiv:2607.05312.

\bibitem[Liu et~al.(2023)Liu, Hu, Ding, and Kong]{Liu2023}
Liu, Y., Hu, Q., Ding, L., and Kong, L. (2023).
Online local differential private quantile inference via self-normalization.
\emph{Proceedings of the 40th International Conference on Machine Learning}, PMLR 202, 21698--21714.

\bibitem[Liu et~al.(2024)Liu, Hu, and Kong]{Liu2024CDF}
Liu, Y., Hu, Q., and Kong, L. (2024).
Tuning-free estimation and inference of cumulative distribution function under local differential privacy.
\emph{Proceedings of the 41st International Conference on Machine Learning}, PMLR 235, 31147--31164.

\begingroup
\bibitem[Massart(1990)]{Massart1990}
Massart, P. (1990).
The tight constant in the Dvoretzky--Kiefer--Wolfowitz inequality.
\emph{Annals of Probability} \textbf{18}, 1269--1283.
\endgroup

\bibitem[Polyak and Juditsky(1992)]{PolyakJuditsky1992}
Polyak, B. T. and Juditsky, A. B. (1992).
Acceleration of stochastic approximation by averaging.
\emph{SIAM Journal on Control and Optimization} \textbf{30}, 838--855.

\bibitem[Robbins and Monro(1951)]{RobbinsMonro1951}
Robbins, H. and Monro, S. (1951).
A stochastic approximation method.
\emph{Annals of Mathematical Statistics} \textbf{22}, 400--407.

\bibitem[Ruppert(1988)]{Ruppert1988}
Ruppert, D. (1988).
Efficient estimators from a slowly convergent Robbins--Monro process.
Technical Report 781, School of Operations Research and Industrial Engineering, Cornell University.

\bibitem[Su and Zhu(2023)]{SuZhu2023}
Su, W. J. and Zhu, Y. (2023).
HiGrad: Uncertainty quantification for online learning and stochastic approximation.
\emph{Journal of Machine Learning Research} \textbf{24}(124), 1--53.

\bibitem[U.S. Census Bureau(2025)]{USCensus2024PUMS}
U.S. Census Bureau (2025).
\emph{2024 ACS 1-Year PUMS User Guide and Overview}.
U.S. Department of Commerce, Washington, DC.
Available at \url{https://www2.census.gov/programs-surveys/acs/tech_docs/pums/2024ACS_PUMS_User_Guide.pdf}.

\bibitem[van der Vaart(1998)]{VanderVaart1998}
van der Vaart, A. W. (1998).
\emph{Asymptotic Statistics}.
Cambridge University Press, Cambridge.

\bibitem[Wang et~al.(2017)Wang, Blocki, Li, and Jha]{Wang2017}
Wang, T., Blocki, J., Li, N., and Jha, S. (2017).
Locally differentially private protocols for frequency estimation.
\emph{26th USENIX Security Symposium}, 729--745.

\bibitem[Warner(1965)]{Warner1965}
Warner, S. L. (1965).
Randomized response: A survey technique for eliminating evasive answer bias.
\emph{Journal of the American Statistical Association} \textbf{60}, 63--69.

\bibitem[Xie et~al.(2025)Xie, Shi, Jiang, Kong, and He]{Xie2025}
Xie, J., Shi, E., Jiang, B., Kong, L., and He, X. (2025).
Online differentially private inference in stochastic gradient descent.
arXiv:2505.08227.

\bibitem[Zhu et~al.(2026)Zhu, Lou, Wei, and Wu]{ZhuLouWeiWu2026}
Zhu, W., Lou, Z., Wei, Z., and Wu, W. B. (2026).
High confidence level inference is almost free using parallel stochastic optimization.
arXiv:2401.09346, version 2.

\end{thebibliography}
\end{document}